\RequirePackage{fix-cm}
\documentclass[twocolumn]{svjour3}          % twocolumn
\smartqed  % flush right qed marks, e.g. at end of proof
\usepackage{graphicx}
\usepackage{amsmath}
\usepackage{amssymb}
\usepackage{mathtools}
\usepackage{tabu}
\usepackage{relsize}
\usepackage{xcolor}
\usepackage{array}
\usepackage[ruled,vlined,commentsnumbered,linesnumbered]{algorithm2e}
\usepackage{bbm}
\usepackage{breakcites}
\usepackage{subfig}   
\usepackage{booktabs}
\usepackage{array,multirow,graphicx}
\usepackage{pifont}
\newcommand{\cmark}{\ding{51}}%
\newcommand{\xmark}{\ding{55}}%

\usepackage[hyphens]{url}
\usepackage[hidelinks]{hyperref}
\hypersetup{breaklinks=true}
\usepackage{apacite}

\newcommand{\postcvpr}[1]{\textcolor{gray}{#1}}
\newcommand{\MPNSeg}{MPNTrackSeg\@\xspace}

\renewcommand{\makeheadbox}{}

\journalname{Arxiv}
\begin{document}
\sloppy

\title{Multi-Object Tracking and Segmentation via Neural Message Passing}
%\subtitle{Subtitle (if any)}


\author{Guillem Brasó* \and Orcun Cetintas* \and Laura Leal-Taixé}


\institute{
*Equal contribution \\
G. Brasó, O. Cetintas, L. Leal-Taixé \at
              Technical University of Munich, Munich, Germany \\
              \email{\{guillem.braso, orcun.cetintas, leal.taixe\}@tum.de} 
}

\date{}

\maketitle

\begin{abstract}

Graphs offer a natural way to formulate Multiple Object Tracking (MOT) and Multiple Object Tracking and Segmentation (MOTS) within the tracking-by-detection paradigm. However, they also introduce a major challenge for learning methods, as defining a model that can operate on such \textit{structured domain} is not trivial. In this work, we exploit the classical network flow formulation of MOT to define a fully differentiable framework based on Message Passing Networks (MPNs). By operating directly on the graph domain, our method can reason globally over an entire set of detections and exploit contextual features. It then jointly predicts both final solutions for the data association problem and segmentation masks for all objects in the scene while exploiting synergies between the two tasks. We achieve state-of-the-art results for both tracking and segmentation in several publicly available datasets. Our code is available at \href{https://github.com/ocetintas/MPNTrackSeg}{\textcolor{magenta}{github.com/ocetintas/MPNTrackSeg}}.

\end{abstract}

%!TEX root = ../1404.tex
\section{Introduction}
 Multiple object tracking (MOT) is the task of determining the bounding box trajectories of all object instances in a video. Multi-object tracking and segmentation (MOTS) \cite{voigtlaender2019mots} extends this task to pixel-level precision by forming trajectories with instance segmentation masks. Both tasks constitute fundamental problems in computer vision, with applications such as autonomous driving, biology, and video analysis.

Segmentation and tracking are naturally intertwined tasks.
When occlusions among objects occur during tracking, masks can disambiguate their relative position in a more explicit manner than boxes.
Even though we can expect improved performance from exploiting the synergies between the tasks, current state-of-the-art barely takes advantage of their interactions.

In recent years, \textit{tracking-by-detection} has been the dominant paradigm among state-of-the-art methods in MOT. This two-step approach consists of first obtaining frame-by-frame object detections and then linking them to form trajectories. While the first task can be addressed with learning-based detectors \cite{faster_rcnn, yolov2}, the latter, data association, is generally formulated as a graph partitioning problem \cite{TangAAS17,yucvpr2007,zhangcvpr2008,lealiccv2011,berclaztpami2011}. In this graph view of MOT, a node represents an object detection, and an edge represents the connection between two nodes. An active edge indicates that two detections belong to the same trajectory. Solving the graph partitioning task, i.e., finding the set of active edges or trajectories, can also be decomposed into two stages. First, a cost is assigned to each edge in the graph encoding the likelihood of two detections belonging to the same trajectory. After that, these costs are used within a graph optimization framework to obtain the optimal graph partition. 

Recent MOTS approaches often design a single model to perform both tasks. A common approach is to employ an additional segmentation module on top of the tracking pipeline and predict masks together with the trajectories \cite{voigtlaender2019mots, porzi2020learning, meinhardt2021trackformer, qiao2021vip}. Despite using a single network, these approaches use separate modules without any explicit interaction to perform segmentation and data association, preventing any synergies between these two tasks from arising. 
An alternative line of work specifically takes advantage of the segmentation masks to extract richer scene information and improve association performance \cite{luiten2020track, xu2020segment}. However, these approaches require an independent segmentation model; hence, they do not simultaneously train for both tasks and therefore cannot adaptively extract segmentation features for data association. All in all, fusing tracking and segmentation cues with a unified model is an under-explored research direction.

We propose to jointly solve data association and segmentation with a unified learning-based solver that can extract and combine relevant appearance, geometry, and segmentation features and reason over the entire scene.
This enables a framework in which association benefits from refined visual information provided by segmentation masks.

With this motivation, we exploit the classical network flow formulation of MOT \cite{zhangcvpr2008} to define our model. Instead of learning pairwise costs, using these within an available solver and independently predicting masks for every object, our method learns to propagate association and segmentation features across the graph in order to directly predict final partitions of the graph together with masks for every object. 
We perform learning directly in the graph domain with a message passing network (MPN) \cite{Gilmer2017NeuralMP}. We design a neural message passing framework that allows our model to learn to combine association and segmentation features into high-order information across the graph. Therefore, we exploit synergies among all tasks in a fully learnable manner while relying on a simple graph formulation.  We show that our unified framework yields substantial improvements with respect to the state-of-the-art both in MOT and MOTS domains without requiring heavily engineered features.

This work builds upon our previous CVPR paper \cite{mpntrack} and extends it by 1) integrating an attentive module to our neural message passing scheme to yield a unified model for multi-object tracking and segmentation and 2) providing an extensive evaluation of our tracking model over three challenging datasets, including MOT20 \cite{mot20}, KITTI \cite{kittiGeiger2012CVPR} and the recently proposed Human in Events dataset \cite{hie}.

To summarize, we make the following {\bf contributions}:
\begin{itemize}
\item We propose a multi-object tracking and segmentation solver based on message passing networks, which can exploit the natural graph structure of the tracking problem to perform both feature learning as well as final solution prediction.
\item We propose a novel time-aware neural message passing update step inspired by classic graph formulations of MOT. 
\item We present a unified framework capable of performing joint tracking and segmentation by combining cues from both domains to improve the association performance. To this end, we propose an attentive message passing update that aggregates inferred temporal and spatial information. 
\item We achieve state-of-the-art results in eight public MOT and MOTS benchmarks.
\end{itemize}

 \section{Related Work}

Most state-of-the-art MOT works follow the tracking-by-detection paradigm which divides the problem into two steps: (i) detecting pedestrian locations independently in each frame, for which neural networks are currently the state-of-the-art~\cite{faster_rcnn,yolov2,sdpdetector}, and (ii) linking corresponding detections across time to form trajectories.

\noindent{\bf Tracking as a Graph Problem.} Data association can be done on a frame-by-frame basis for online applications \cite{breitensteiniccv2009, esscvpr2008, pellegriniiccv2009} or track-by-track \cite{berclazcvpr2006}. For video analysis tasks that can be done offline, batch methods are preferred since they are more robust to occlusions.
The standard way to model data association is by using a graph, where each detection is a node, and edges indicate possible links among them. The data association can then be formulated as a maximum flow
\cite{berclaztpami2011} or, equivalently, a minimum cost problem with either fixed costs based on distance \cite{jiangcvpr2007,pirsiavashcvpr2011,zhangcvpr2008}, including motion models \cite{lealiccv2011}, or learned costs \cite{lealcvpr2014}. Both formulations can be solved optimally and efficiently. Alternative formulations typically lead to more involved optimization
problems, including minimum cliques~\cite{zamireccv2012} and lifted disjoint paths ~\cite{pmlr-v119-hornakova20a, aplift},
general-purpose solvers, e.g., multi-cuts~\cite{TangAAS17}. 
A recent trend is to design ever more complex models which include other vision input such as reconstruction for multi-camera sequences \cite{lealcvpr2012,wucvpr2011}, activity recognition \cite{choieccv2012}, segmentation \cite{milancvpr2015}, keypoint trajectories \cite{choiiccv2015} or joint detection \cite{TangAAS17}.

\noindent{\bf Learning in Graph-based Tracking.}
It is no secret that neural networks are now dominating the state-of-the-art in many vision tasks since \cite{krizhevskyImageNet} showed their potential for image classification. The trend has also arrived in the tracking community, where learning has been used primarily to learn a mapping from image to optimal costs for the aforementioned graph algorithms. The authors of \cite{lealcvprw2016} use a siamese network to directly learn the costs between a pair of detections, while a mixture of CNNs and recurrent neural networks (RNN) is used for the same purpose in \cite{Sadeghian_2017_ICCV, longterm_tracklet}. The authors of \cite{ristanicvpr2018} show the importance of learned ReIdentification (ReID)  features for multi-object tracking. More recently, JDE and FairMOT\cite{towards_realtime_mot, fairmot} explore the potential of jointly learning appearance features data association and detection, therefore, improving efficiency.
All aforementioned methods learn the costs independently from the optimization method that actually computes the final trajectories. In contrast, \cite{kimaccv12,Wang2015b,Schulter_2017_CVPR} incorporate the optimization solvers into learning. The main idea behind these methods is that costs also need to be optimized for the solver in which they will be used. \cite{kimaccv12,Wang2015b, end_to_end_urtasun} rely on structured learning losses while \cite{Schulter_2017_CVPR} proposes a more general bi-level optimization framework. These works can be seen as similar to ours in spirit, given our common goal of incorporating the full inference model into learning for MOT. However, we follow a different approach towards this end: we propose to directly learn a solver and treat data association as a classification task, while their goal is to adapt their methods to perform well with non-learnable solvers. Moreover, all these works are limited to learning either pairwise costs \cite{end_to_end_urtasun, Schulter_2017_CVPR} or additional quadratic terms \cite{Wang2015b, kimaccv12} but cannot incorporate higher-order information as our method. Instead, we propose to leverage the common graph formulation of MOT as a domain in which to perform learning. Concurrent work also explores the potential of learning in the graph domain \cite{gsm, graph_nets_tracking}; however, they do so within frame-by-frame settings, while we focus on a more general graph formulation. Moreover, our proposed approach has inspired recent graph neural network-based work to tackle proposal classification for Multiple Hypothesis Tracking \cite{lpcmot} and graph-matching-based tracking \cite{graph_matching}.

\noindent{\bf Regression-Based Tracking.} 
 While graph-based methods are still actively used by several state-of-the-art trackers, several recent works achieve remarkable performance with a simpler formulation. The first of such works was Tracktor  ~\cite{tracktor} which performs tracking by exploiting the regressor head of a Faster R-CNN to sequentially predict the location objects in consecutive frames. TMOH~\cite{tmoh} improves Tracktor's performance under occlusion, and CTracker~\cite{ctracker} builds upon its idea with a framework that performs chained regression over consecutive pairs of frames and uses attention.  In a similar fashion to these works,
 CenterTrack ~\cite{10.1007/978-3-030-58548-8_28} uses center-points instead of bounding boxes and predicts offset heatmaps to regress consecutive object locations, and PermaTrack\cite{permatrack} improves its robustness to occlusions by predicting offsets over multiple future frames. In our approach, we leverage Tracktor as an initial preprocessing step to improve object detections before applying our neural solver over the result.
\\
\noindent{\bf Deep Learning on Graphs.} 
Graph Neural Networks (GNNs) were first introduced in \cite{scarselli_graph_nn} as a generalization of neural networks that can operate on graph-structured domains. Since then, several works have focused on further developing and extending them by developing convolutional variants  \cite{Bruna2013SpectralNA, deferrand_cnns, kipf2016semi}. More recently, most methods were encompassed within a more general framework termed {neural message passing} \cite{Gilmer2017NeuralMP} and further extended in \cite{battaglia_graph_networks} as {graph networks}. Given a graph with some initial features for nodes and optionally edges, the main idea behind these models is to embed nodes (and edges) into representations that take into account not only the node's features but also those of its neighbors in the graph, as well as the overall graph topology. These methods show remarkable performance in a wide variety of areas, ranging from chemistry \cite{Gilmer2017NeuralMP} to combinatorial optimization \cite{combinatorial_cnns}. Within vision, they have been successfully applied to problems such as human action recognition \cite{graph_nets_action_recognition}, visual question answering \cite{graph_nets_vqa} or single object tracking \cite{Gao_2019_CVPR}.

\noindent{\bf Multi-Object Tracking and Segmentation.} MOTS \cite{voigtlaender2019mots} was recently introduced as an extension of MOT to overcome the limitations of bounding boxes while also opening up the possibility to incorporate a richer visual cue for the association step. Prior works often solve both tasks by modifying strong segmentation models and extending them with an association head \cite{voigtlaender2019mots, porzi2020learning, qiao2021vip, Wu_2021_CVPR}. TrackR-CNN \cite{voigtlaender2019mots} integrates temporal information into the MaskR-CNN framework with 3D convolutions, while MOTSNet~\cite{porzi2020learning} employs a mask-pooling layer for filtering out the background information from the instance feature maps. Alternatively, recent works attempt to condition the association step on masks by first performing segmentation and then reconstructing objects in the 3D space \cite{luiten2020track} or denoting objects with 2D point clouds~\cite{xu2020segment}. The motivation of this line of work is to utilize the strong visual cues provided by masks to improve the association performance of the model. However, these approaches disentangle the mask prediction step from the association. Thus, simultaneously solving both tasks by fusing tracking and segmentation cues remains an open question.

\begin{figure*}[h]
\vspace{-0.2cm}
        \centering
           \subfloat[Input]{%
              \includegraphics[height=5.5cm]{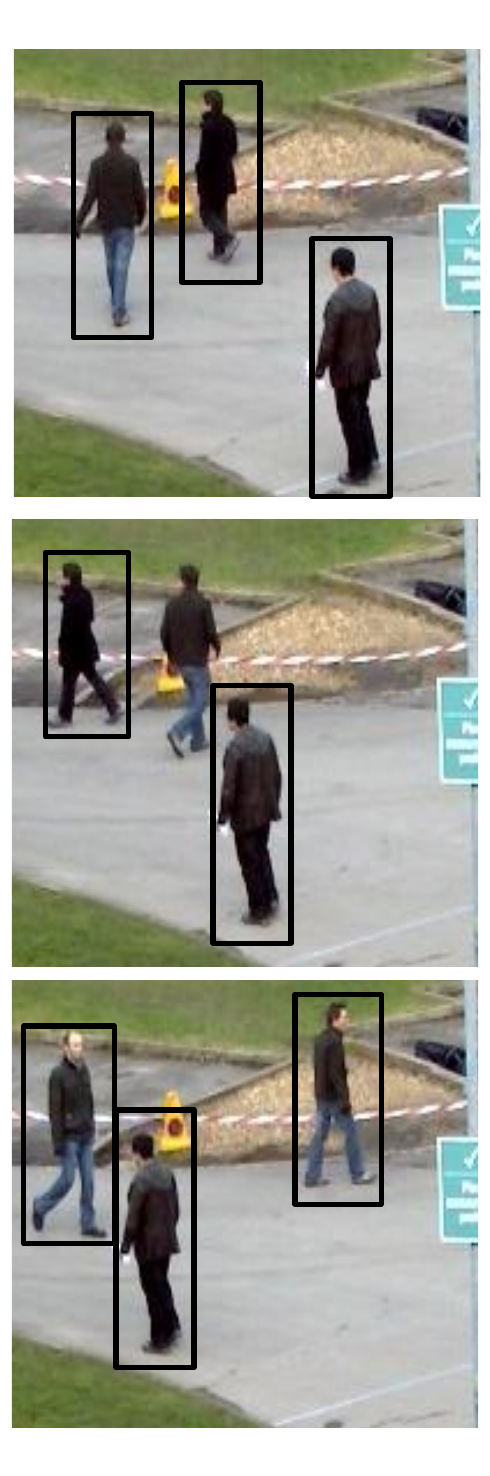}
              \label{input_fig}
           } 
           \hspace{0.3cm}
           \subfloat[Graph Construction + Feature Encoding]{%
              \includegraphics[height=5.5cm]{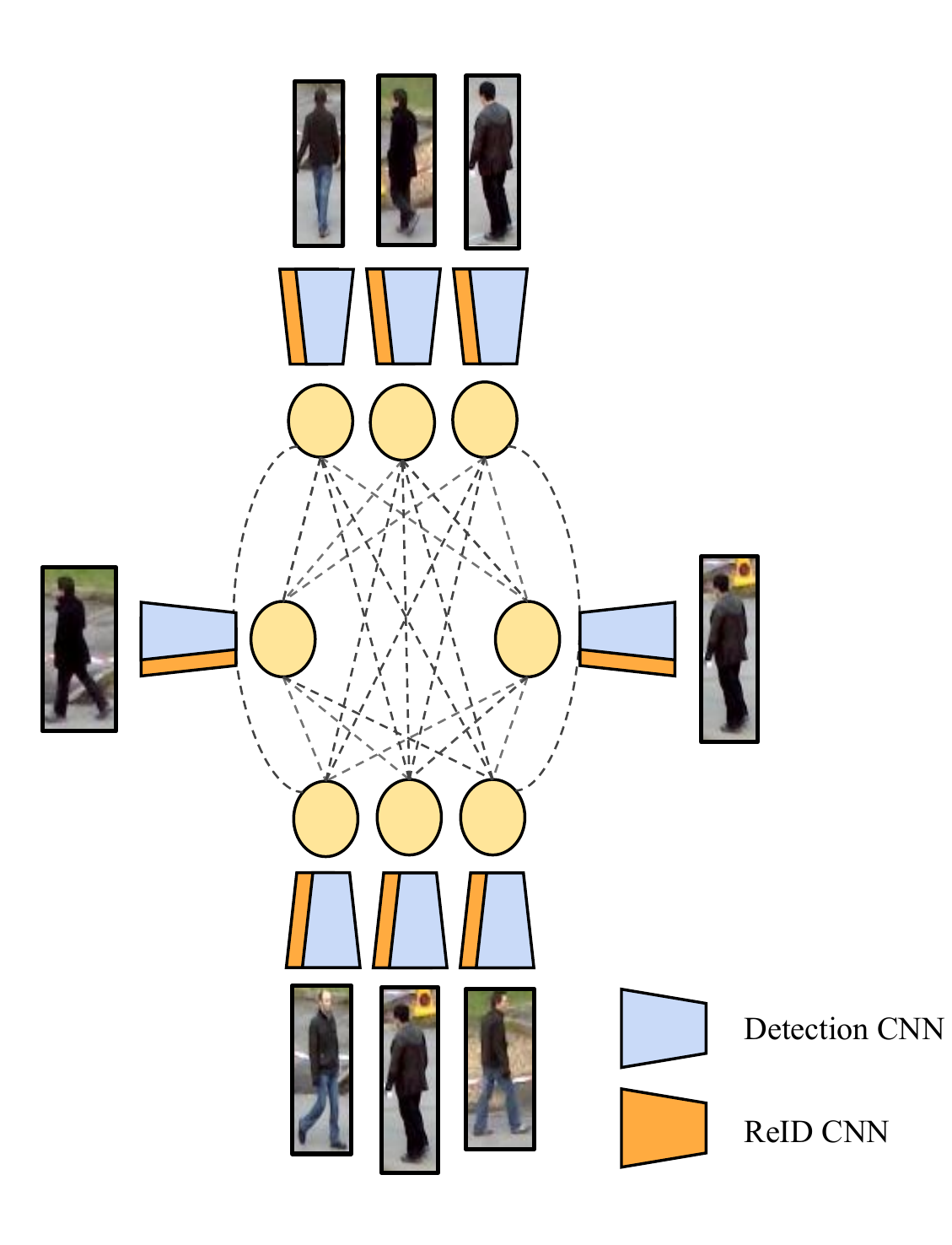}
              \label{graph_init}
           }
          \hspace{0.3cm}
           \subfloat[Neural Message Passing]{%
              \includegraphics[height=5.5cm]{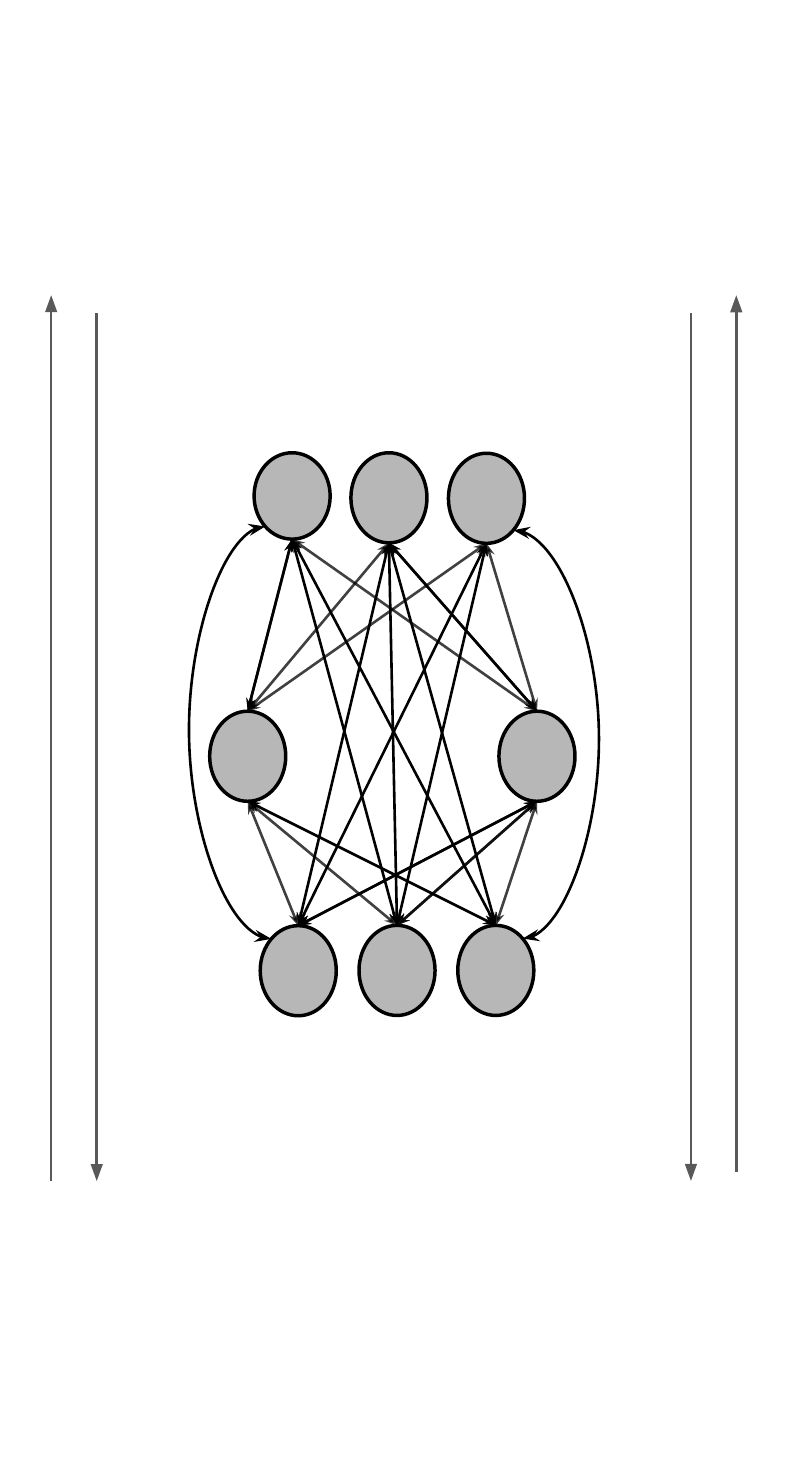} 
              \label{mp_pipeline}
           }
          \hspace{0.3cm}
           \subfloat[Edge Classification + Mask Prediction]{%
              \includegraphics[height=5.5cm]{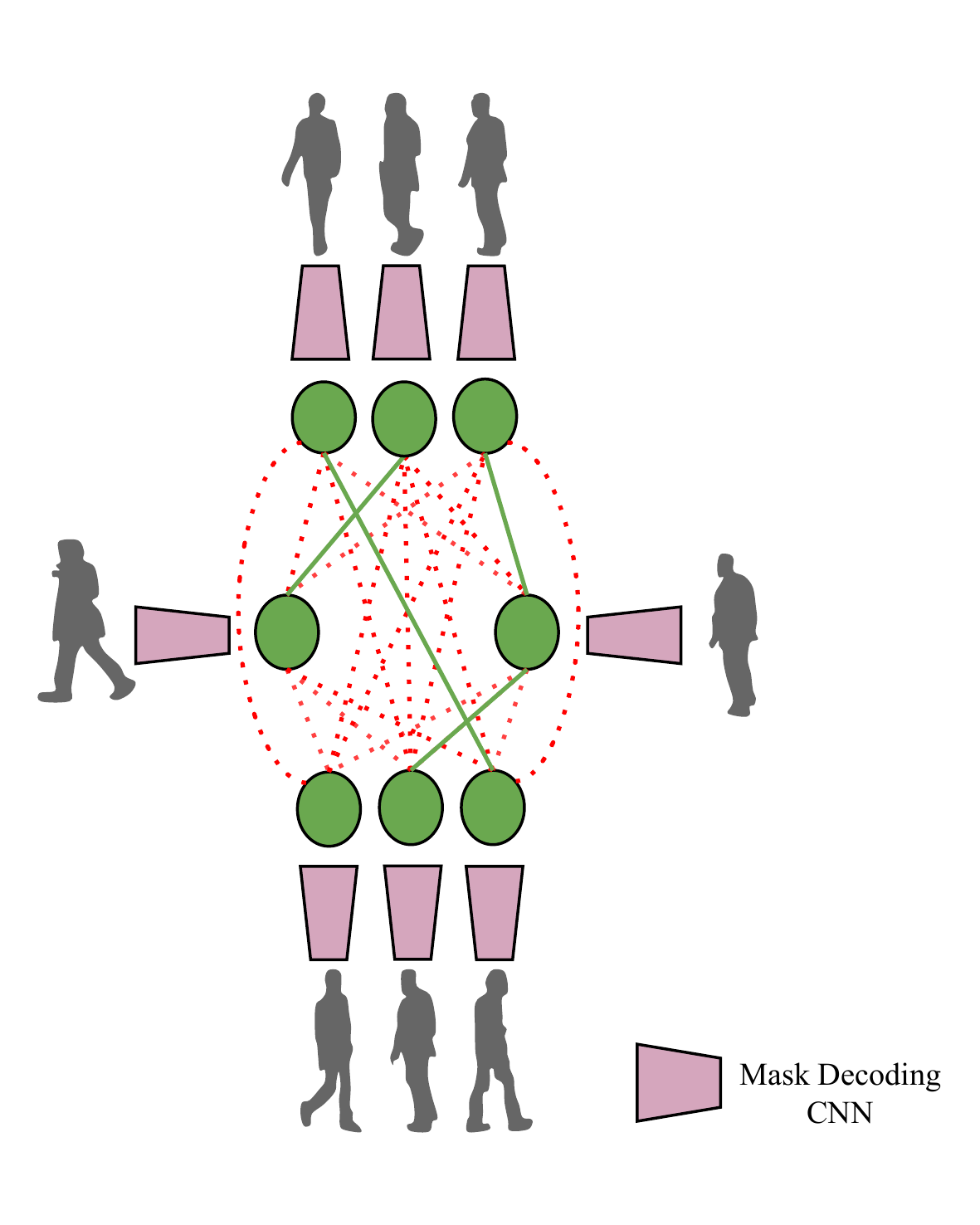} \label{edge_class_pipeline}
           }
          \hspace{0.3cm}
           \subfloat[Output]{%
              \includegraphics[height=5.5cm]{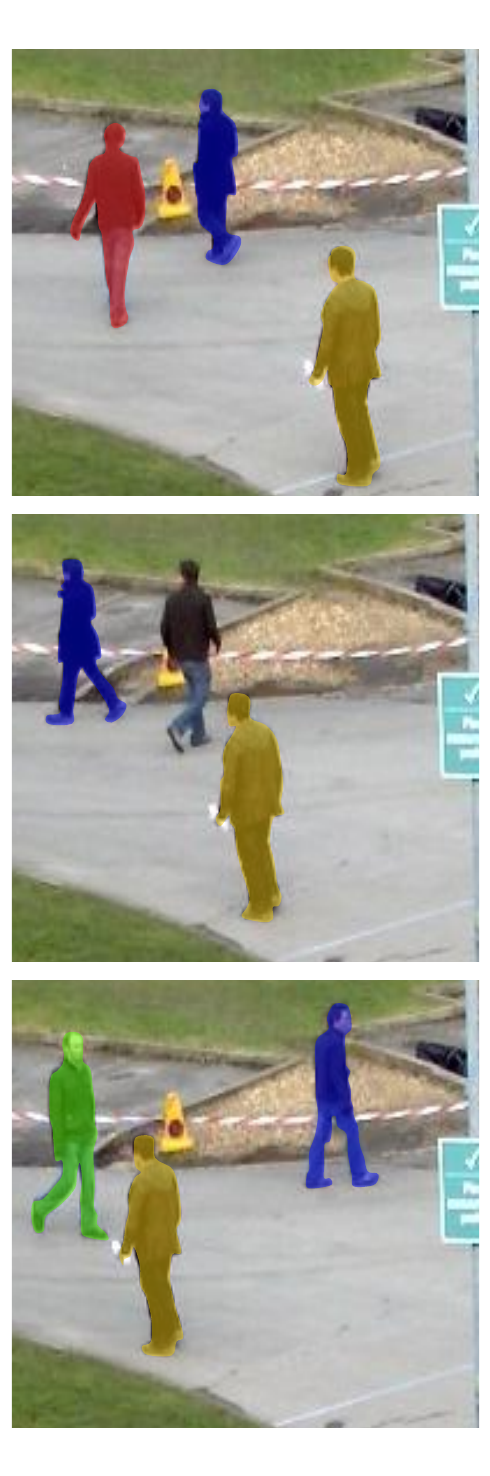} \label{output_pipeline}
           }           
\vspace{-0.2cm}
           \caption{Overview of our method. (a) We receive as input a set of frames and detections. (b) We construct a graph in which nodes represent detections, and all nodes at different frames are connected by an edge. (c) We initialize node embeddings in the graph with two CNNs that encode appearance and mask features. Edge embeddings are initialized with an MLP encoding geometry information (not shown in the figure). (c) The information contained in these embeddings is propagated across the graph for a fixed number of iterations through neural message passing. (d) Once this process terminates, the embeddings resulting from neural message passing are used to predict masks and classify edges into active (colored with green) and non-active (colored with red). During training, we compute the cross-entropy loss of our predictions w.r.t. ground truth labels. (e) At inference, we follow a simple rounding scheme to binarize our classification scores and obtain final trajectories.}
           \label{pipeline}
\end{figure*}

\section{Tracking as a Graph Problem} \label{problem_formulation}

Our method's tracking formulation is based on the classical min-cost flow view of MOT \cite{zhangcvpr2008}. In order to provide some background and formally introduce our approach, we start by providing an overview of the network flow MOT formulation. We then explain how to leverage this framework to reformulate the data association task as a learning problem.

\subsection{Problem Statement}
In tracking-by-detection, we are given as input a set of object detections $\mathcal{O}=\{o_1, \dots , o_n\}$, where $n$ is the total number of objects for all frames of a video. Each detection is represented by $o_i = (a_i, p_i,t_i)$, where $a_i$ denotes the raw pixels of the bounding box, $p_i$ contains its 2D image coordinates and $t_i$ its timestamp. A trajectory is defined as a set of time-ordered object detections $T_i=\{{ o_{i_1}}, \dots , { o_{i_{n_i}}}\}$, where $n_i$ is the number of detections that form trajectory $i$. The goal of MOT is to find the set of trajectories $\mathcal{T}_*=\{T_1, \dots, T_m\}$, that best explains the observations $\mathcal{O}$. 
The problem can be modelled with an undirected graph $G=(V, E)$, where $V:=\{1, \dots ,n\}$, $E \subset V \times V $, and each node $i \in V$ represents a unique detection $o_i \in \mathcal{O}$. The set of edges $E$ is constructed so that every pair of detections, i.e., nodes, in different frames is connected, hence allowing to recover trajectories with missed detections. Now, the task of dividing the set of original detections into trajectories can be viewed as grouping nodes in this graph into disconnected components. Thus, each trajectory $T_i=\{{ o_{i_1}}, \dots , { o_{i_{n_i}}}\}$ in the scene can be mapped into a group of nodes  $\{i_1, \dots , i_{n_i}\}$ in the graph and vice-versa.

\subsection{Network Flow Formulation} 
\label{sec:netflow}
In order to represent graph partitions, we introduce a binary variable for each edge in the graph. In the classical minimum cost flow formulation~\cite{zhangcvpr2008}, this label is defined to be 1 between edges connecting nodes that (i) belong to the same trajectory, and (ii) are temporally consecutive inside a trajectory; and 0 for all remaining edges. 

A trajectory $T_i=\{{ o_{i_1}}, \dots , { o_{i_{n_i}}}\}$ is equivalently denoted by the set of edges $\{(i_1, i_2), \dots ,(i_{n_i - 1}, i_{n_i})\} \subset E$, corresponding to its time-ordered path in the graph. We will use this observation to formally define the edge labels. For every pair of nodes in different timestamps, $(i, j) \in E$, we define a binary variable $y_{(i, j)}$ as:
$$
     y_{(i, j)} \coloneqq \begin{cases}
               1               & \exists T_k \in \mathcal{T}_* \text{ s.t. } (i, j) \in T_k \\
               0               & \text{otherwise.}
           \end{cases}
$$
An edge $(i,j)$ is said to be \textit{active} whenever $y_{(i, j)}=1$. 
We assume trajectories in $\mathcal{T}$ to be node-disjoint, i.e., a node cannot belong to more than one trajectory. Therefore, $y$ must satisfy a set of linear constraints. For each node $i \in V$: 

\begin{align}
& &\sum_{(j, i)\in E \text{ s.t. } t_i>t_j} y_{(j, i)} &\leq 1  \label{flow_in_constr} \\
& &\sum_{(i, k)\in E \text{ s.t. } t_i<t_k} y_{(i, k)} &\leq 1\label{flow_out_constr}
\end{align}

These inequalities are a simplified version of the \textit{flow conservation constraints}~\cite{networkflows}. In our setting, they enforce that every node gets linked via an active edge to, at most, one node in past frames and one node in upcoming frames.

\subsection{From Learning Costs to Predicting Solutions}
In order to obtain a graph partition with the framework we have described, the standard approach is to first associate a cost $c_{(i, j)}$ to each binary variable $y_{(i, j)}$. This cost encodes the likelihood of the edge being active~\cite{lealcvpr2014, lealcvprw2016, Schulter_2017_CVPR}. The final partition is found by optimizing:
\begin{equation*}
\begin{array}{ll@{}ll}
&\min_{y}  & \mathlarger{ \sum }_{(i, j) \in E} {c_{(i, j)}y_{(i, j)}}     &\\
\text{Subject to:} & \text{Equation } (\ref{flow_in_constr}) &\\
                &  \text{Equation } (\ref{flow_out_constr})      &\\                                               
&y_{(i,j)} \in \{0,1\}, & \quad (i, j) \in E
\end{array}
\end{equation*}
which can be solved with available solvers in polynomial time~\cite{berclazcvpr2006, networkflows}.

We propose to, instead, directly learn to predict which edges in the graph will be active, i.e., predict the final value of the binary variable $y$. 
To do so, we treat the task as a classification problem over edges, where our labels are the binary variables $y$. Overall, we exploit the classical network flow formulation we have just presented to treat the MOT problem as a fully learnable task.

\section{Learning to Track with Message Passing Networks}
Our main contribution is to exploit the graph formulation described in the previous section to design a differentiable framework for joint tracking and segmentation. Given a set of detections, we design a neural message passing network that operates on its underlying graph and extracts contextual node and edge embeddings. We classify each edge embedding to predict the values of the binary \textit{flow} variables $y$ directly, and we exploit node embeddings to obtain a segmentation mask for every target.  Our method is based on a novel message passing network (MPN) and is able to capture the graph structure of the MOT and MOTS problems. Within our proposed MPN framework, appearance, geometry, and segmentation cues are propagated across the entire set of detections, allowing our model to reason globally about the entire graph.

\subsection{Overview}
Our method consists of the following four main stages:
\noindent{\bf 1. Graph Construction:} Given a set of object detections in a video, we construct a graph where nodes correspond to detections and edges correspond to connections between nodes (Section \ref{sec:netflow}).

\noindent{\bf 2. Feature Encoding:} We initialize the node feature embeddings from two convolutional neural networks (CNNs) applied on every detection's Region of Interest (RoI), extracting appearance and mask features. For each edge, i.e., for every pair of detections in different frames, we compute a vector with features encoding their bounding box relative size, position, and time distance. We then feed it to a multi-layer perceptron (MLP) that returns a {\it geometry} embedding (Section \ref{feature_encoding}).

\noindent{\bf 3. Neural Message Passing:} We perform a series of message passing steps over the graph. Intuitively, for each round of message passing, nodes share appearance information with their connecting edges, and edges share geometric information with their incident nodes. This yields updated embeddings for nodes and edges containing {\it higher-order} information that depends on the overall graph structure (Section \ref{vanilla_mpns}, \ref{time_aware_message_passing} and \ref{att_message_passing}).

\noindent{\bf 4. Classification:} We use the final edge embeddings to perform binary classification into active/non-active edges and node embeddings to classify each pixel of the RoIs into foreground/background. We train our model using the cross-entropy loss for both tasks (Section \ref{inference}).

At test time, we use our model's prediction per edge as a continuous approximation (between 0 and 1) of the target \textit{flow} variables. We then follow a simple scheme to round them and obtain the final trajectories. For a visual overview of our pipeline, see Figure \ref{pipeline}.

\begin{figure*}[h]
        \centering
           \subfloat[Initial Setting]{%
              \includegraphics[height=6cm]{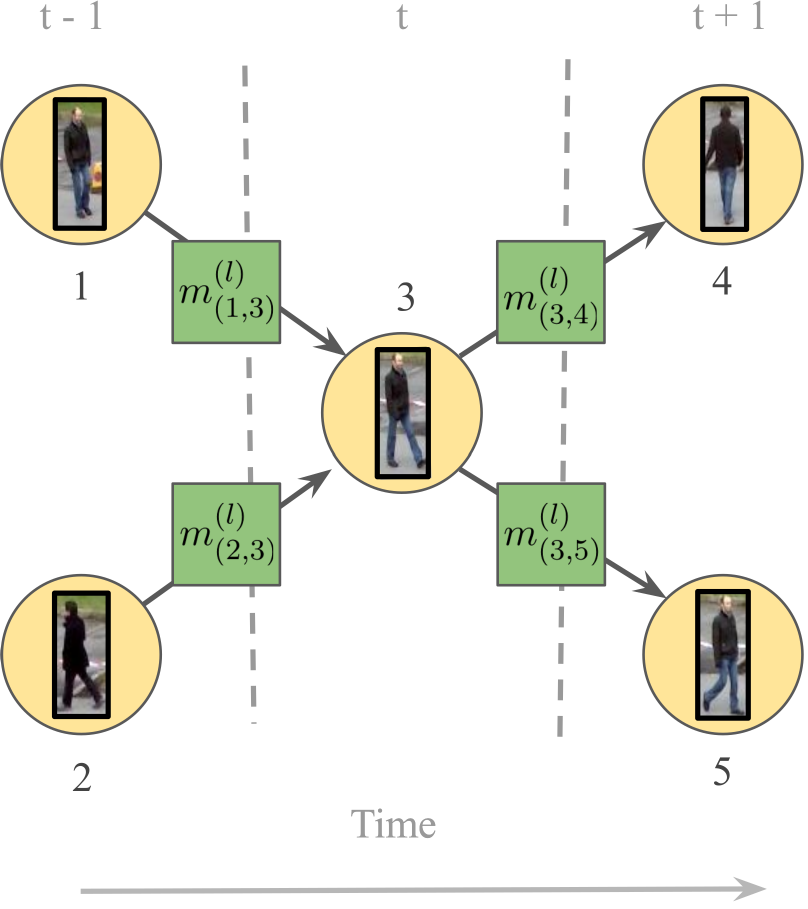}
              \label{setting}
           } 
           \hspace{0.3cm}
           \subfloat[Vanilla node update]{%
              \includegraphics[height=6cm]{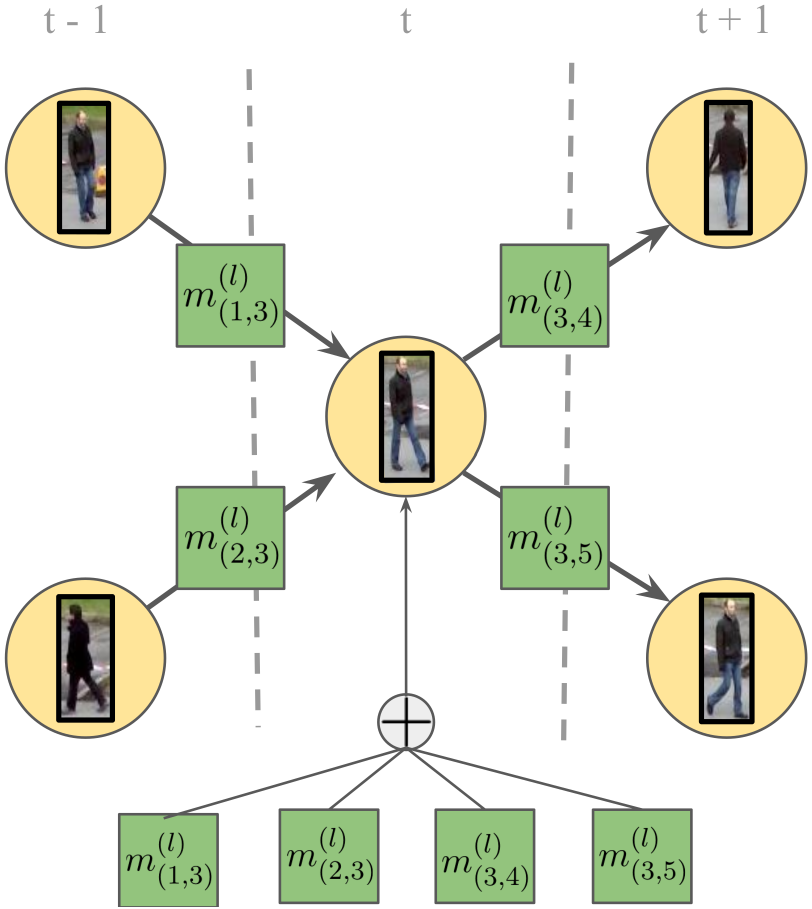}
              \label{vanilla_update}
           }
          \hspace{0.3cm}
           \subfloat[Time-aware node update]{%
              \includegraphics[height=6cm]{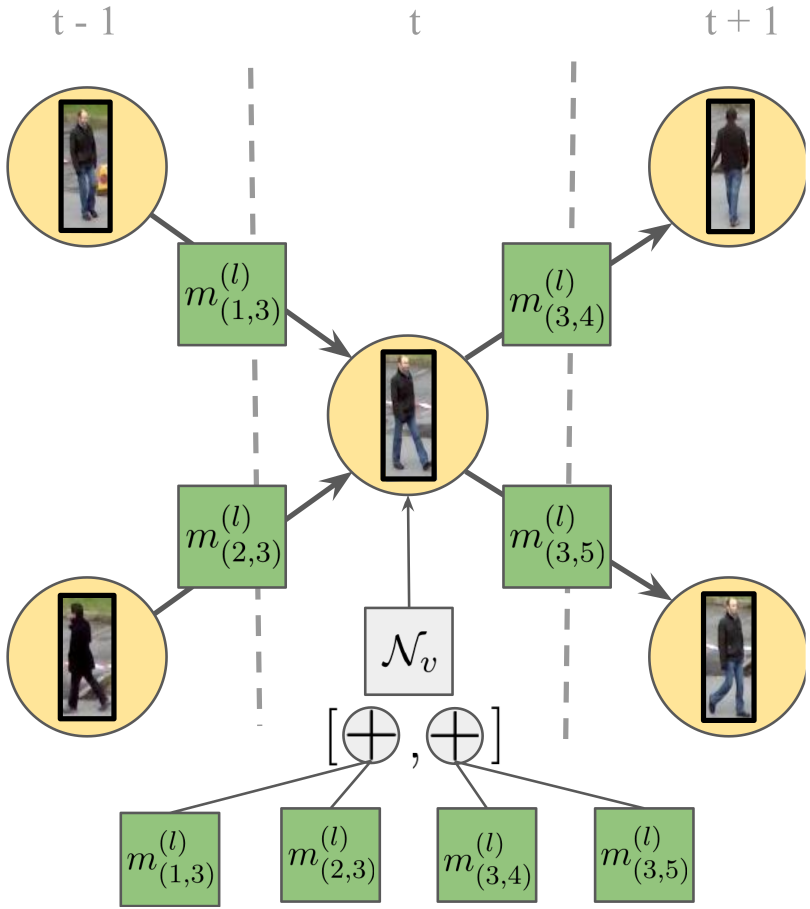} \label{time_aware_update}
           }
           \caption{Visualization of node updates during message passing. Arrow directions in edges show time direction. Note the time division in $t-1$, $t$, and $t+1$. In this case, we have $N_3^{past}=\{1, 2\}$ and $N_3^{fut}=\{4, 5 \}$. \ref{setting} shows the starting point after an edge update has been performed (equation \ref{node2edge}), and the intermediate node update embeddings (equation \ref{edge2node1}) have been computed. \ref{vanilla_update} shows the standard node update in vanilla MPNs, in which all neighbors' embeddings are aggregated jointly. \ref{time_aware_update} shows our proposed update, in which embeddings from past and future frames are aggregated separately, then concatenated and fed into an MLP to obtain the new node embedding.}\label{node_updates}
           \label{fig:default}
\end{figure*}      

\subsection{Message Passing Networks} \label{vanilla_mpns}
In this section, we provide a brief introduction to MPNs based on the work presented in \cite{Gilmer2017NeuralMP, kipf_icml2018, interaction_nets_battaglia, battaglia_graph_networks}. Let $G = (V, E)$ be a graph. Let $h_i^{(0)}$ be a node embedding for every $i\in V$, and $h_{(i, j)}^{(0)}$ an edge embedding for every $(i, j)\in E$. The goal of MPNs is to {learn} a function to propagate the information contained in nodes and edge feature vectors across $G$.
The propagation procedure is organized in {embedding updates} for edges and nodes, which are known as {\it message passing steps}~\cite{Gilmer2017NeuralMP}. In \cite{battaglia_graph_networks, kipf_icml2018, interaction_nets_battaglia}, each message passing step is divided, in turn, into two updates: one from nodes to edges $(v\rightarrow e)$, and one from edges to nodes $(e\rightarrow v)$. 
The updates are performed sequentially for a fixed number of iterations $L$. For each $l\in \{1, \dots, L\}$, the general form of the updates is the following \cite{battaglia_graph_networks}:
\begin{align}
&(v\rightarrow e) &h_{(i, j)}^{(l)} &= \mathcal{N}_e\left([h_i^{(l-1)}, h_j^{(l-1)}, h_{(i, j)}^{(l-1)}]\right)  \label{node2edge}\\
&(e\rightarrow v) 
            &m_{(i, j)}^{(l)} &=  \mathcal{N}_v\left([h_i^{(l-1)}, h_{(i, j)}^{(l)} ]\right)  \label{edge2node1} \\
&            &h_i^{(l)} &= \Phi \left(\left \{ m_{(i, j)}^{(l)}   \right\}_{ j \in N_i}\right)  \label{edge2node2} 
\end{align}
$\mathcal{N}_e$ and $\mathcal{N}_v$ represent learnable functions, e.g., MLPs, that are shared across the entire graph. $[.]$ denotes concatenation, $N_i\subset V$ is the set of adjacent nodes to $i$, and $\Phi$ denotes an order-invariant operation, e.g., a summation, maximum, or an average. 
Note, after $L$ iterations, each node contains information about all other nodes at a distance $L$ in the graph. Hence, $L$ plays an analogous role to the {receptive field} of CNNs, allowing embeddings to capture context information.

\subsection{Time-Aware Message Passing} \label{time_aware_message_passing}
The previous message passing framework was designed to work on arbitrary graphs. 
However, MOT graphs have a very specific structure that we propose to exploit. Our goal is to encode a MOT-specific inductive bias in our network, specifically in the node update step.
Recall the node update depicted in Equations~\ref{edge2node1} and \ref{edge2node2}, which allows each node to be compared with its neighbors and aggregate information from all of them to update its embedding with further context. 
Recall also the structure of our flow conservation constraints (Equations~\ref{flow_in_constr} and \ref{flow_out_constr}), which imply that each node can be connected to, at most, one node in future frames and another one in past frames. Arguably, aggregating all neighboring embeddings at once makes it difficult for the updated node embedding to capture whether these constraints are being violated or not (see Section \ref{ablation_study} for constraint satisfaction analysis). 
More generally, explicitly \textit{encoding} the temporal structure of MOT graphs into our MPN formulation can be a useful prior for our learning task. 
Towards this goal, we modify Equations~\ref{edge2node1} and \ref{edge2node2} into time-aware update rules by dissecting the aggregation into two parts: one over nodes in the past, and another over nodes in the future. 
Formally, let us denote the neighboring nodes of $i$ in future and past frames by ${N}^{fut}_i$ and ${N}^{past}_i$, respectively. 
Let us also define two different MLPs, namely, $\mathcal{N}^{fut}_v$ and $\mathcal{N}^{past}_v$. 
At each message passing step $l$ and for every node $i \in V$, we start by computing \textit{past} and \textit{future} edge-to-node embeddings for all of its neighbors $j\in {N}_i$ as:
%\vspace{-0.2cm}
\begin{align}
&    &m_{(i, j)}^{(l)} &= \begin{cases} \mathcal{N}^{past}_v\left([h_i^{(l-1)}, h_{(i, j)}^{(l)}, h_{(i)}^{(0)}]\right)  \text{ if } &j\in N^{past}_i \\
                  \mathcal{N}^{fut}_v \text{ }\left([h_i^{(l-1)}, h_{(i, j)}^{(l)}, h_{(i)}^{(0)}]\right) \text{   if } &j\in N^{fut}_i 
           \end{cases}
\label{eq:m_case}
\end{align}
%\vspace{-0.2cm}
Note, the initial embeddings $h_{(i)}^{(0)}$ have been added to the computation. This skip connection ensures that our model does not \textit{forget} its initial features during message passing, and we apply it analogously with initial edge features in Equation~\ref{node2edge}. After that, we aggregate these embeddings separately, depending on whether they were in future or past positions with respect to $i$:
\begin{align}
&    &h_{i, past}^{(l)} &= \sum_{ j \in N^{past}_i} m^{(l)}_{(i, j)} \\
&    &h_{i, fut}^{(l)} &= \sum_{ j \in N^{fut}_i} m^{(l)}_{(i, j)} 
\label{eq:h_past_future}
\end{align}
Now, these operations yield \textit{past} and \textit{future} embeddings $h_{i, past}^{(l)}$ and $h_{i, fut}^{(l)}$, respectively. We compute the final updated node embedding by concatenating them and feeding the result to one last MLP, denoted as $\mathcal{N}_v$:
\begin{align}
&    &h_i^{(l)} &=  \mathcal{N}_v([h_{i, past}^{(l)} , h_{i, fut}^{(l)} ])
\label{eq:h}
\end{align}
We summarize  our time-aware update in Figure~\ref{time_aware_update}. As we demonstrate experimentally (see Section \ref{ablation_study}), this simple architectural design results in a significant performance improvement with respect to the \textit{vanilla} node update of MPNs, shown in Figure~\ref{vanilla_update}.

\subsection{Attentive Message Passing} 
\label{att_message_passing}

Our time-aware message passing framework utilizes appearance and geometry feature vectors of dimension $d$ for the association. Segmentation, on the other hand, is a dense prediction task that requires pixel-precise outputs. Preserving spatial information is crucial for this task, therefore we incorporate contextual mask features with additional spatial dimensions $H$ and $W$ (i.e., tensors in $\mathbb{R}^{H\times W \times d'}$) for every node into our message passing updates. $H$ and $W$ correspond, respectively, to the height and width at which each node's RoI is resized.

Our goal is to extract rich segmentation features encoding temporal information to be used for association. Towards this end, we leverage the temporal information encoded in the edge embeddings and use them to produce attention coefficients to guide the updates of mask features. The mask features will therefore be most influenced by neighbors in the graph belonging to the same trajectory.

Formally, let $\tilde{h}_i^{(0)} \in \mathbb{R}^{H\times W \times d}$ represent a secondary mask node embedding for every $i\in V$ encoding visual information. In addition, let $\mathcal{N}_{e}^{w}$ denote an MLP working on the edges of the graph. At each message passing step $l\geq1$, we calculate the unnormalized attention weights for each edge of the graph:

\begin{align}
&  &w_{(i, j)}^{(l)} &= \mathcal{N}_{e}^{w}\left(h_{(i,j)}^{(l-1)}\right)
\label{eq:3d_logits}
\end{align}

The attention scores for the edge $(i, j)$ are computed by respecting the time-aware message passing rules and normalizing the weights over the \textit{past} and \textit{future} neighbors of the node $i$ with a softmax operator:

\begin{align}
& &a_{(i, j)}^{(l)} &= \begin{cases} 
\frac{\exp\left(w_{(i, j)}^{(l)}\right)}{\sum\limits_{ k \in N^{past}_i}\exp\left(w_{(i, k)}^{(l)}\right)} &\text{ if } \quad j\in N^{past}_i 
 \\
\frac{\exp\left(w_{(i, j)}^{(l)}\right)}{\sum\limits_{ k \in N^{fut}_i}\exp\left(w_{(i, k)}^{(l)}\right)} &\text{ if } \quad j\in N^{fut}_i 
 \end{cases}
\end{align}

The \textit{past} and \textit{future} context tensors for each node $i$ are, then, computed as a weighted sum of the node features of its neighbors:

\begin{align}
&    &c_{i, past}^{(l)} &= \sum_{ j \in N^{past}_i} a_{(i, j)}^{(l)} \tilde{h}^{(l-1)}_{j} \\
&    &c_{i, fut}^{(l)} &= \sum_{ j \in N^{fut}_i} a_{(i, j)}^{(l)} \tilde{h}^{(l-1)}_{j} 
\end{align}

Finally, we obtain the updated node embeddings by concatenating initial embedding $\tilde{h}_i^{(0)}$ with the \textit{past} and \textit{future} context tensors and feeding the result to a 2-layer CNN, denoted as $\tilde{\mathcal{N}}_v$:

\begin{align}
&    &\tilde{h}_{i}^{(l)} &=  \tilde{\mathcal{N}}_v([c_{i, past}^{(l)} , c_{i, fut}^{(l)}, \tilde{h}_{(i)}^{(0)}])
\end{align}

Note that high-dimensional secondary node embeddings $\tilde{h}_i$ and attentive update rules are only employed in the MOTS setting. Default node embeddings $h_i$ are still present in this scenario and are updated according to Equations~\ref{eq:m_case}-\ref{eq:h}.

\begin{figure*}[h]
        \centering
            \includegraphics[height=6cm]{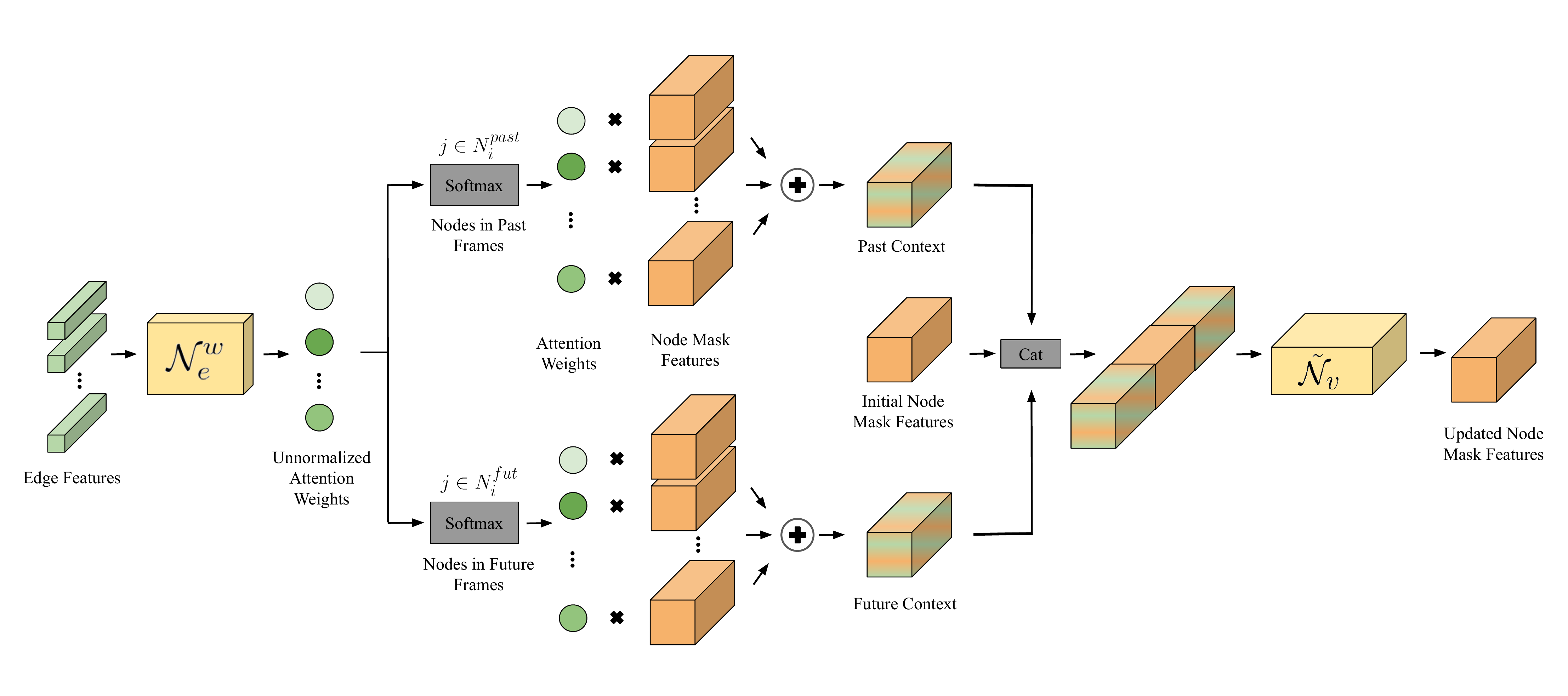}
           \caption{Visualization of our proposed attentive message passing. Updates of the node mask features (orange) are guided by the attention weights, obtained from the edge features (green). }
           \label{3d_update}
\end{figure*}

We illustrate our update scheme in Figure \ref{3d_update}. With our attentive message passing updates
high-dimensional mask features can participate in the message passing steps, and we can train our model for tracking and segmentation jointly. It is worth highlighting that updates of these node features are directly governed by the edges of the graph. Hence, in our end-to-end differentiable framework, both the edge features $h_{(i, j)}^{(l)}$ (later classified for tracking) and node features  $\tilde{h}_i^{(l)}$ (later used for segmentation) are guided by both the tracking and segmentation objectives. Overall, we obtain a joint framework in which segmentation features can guide association decisions. In Section~\ref{ablation_study}, we empirically show that our unified pipeline improves upon our baseline in which segmentation and tracking are detached.

\subsection{Feature Encoding} \label{feature_encoding}

The initial embeddings that our MPN receives as input are produced by other backpropagatable networks.

\noindent{\bf Appearance and Mask Embeddings.} We rely on a CNN, denoted as  $\mathcal{N}_v^{enc}$, to \textit{learn} to extract feature embeddings directly from RGB data. For every detection $o_i \in \mathcal{O}$, and its corresponding image patch $a_i$, we obtain $o_i$'s corresponding node embedding by computing $h_i^{(0)}:=\mathcal{N}_v^{enc}(a_i)$. Moreover, we use an additional network composed of a CNN backbone and RoI Align \cite{He_2017_ICCV} denoted as $\tilde{\mathcal{N}}_v^{enc}$ to obtain secondary node embeddings from each Region of Interest (RoI) $r_i$ as $\tilde{h}_i^{(0)}:=\mathcal{\tilde{N}}_v^{enc}(r_i)$.

\noindent{\bf Geometry Embedding.} For each pair of detections in different frames, we seek to obtain a representation that encodes their relative position, size, as well as distance in time. For every pair of detections $o_i$ and $o_j$ with timestamps $t_i \neq t_j$, we consider their bounding box coordinates parameterized by top left corner image coordinates, height and width, i.e., $(x_i, y_i, h_i, w_i)$ and $(x_j, y_j, h_j, w_j)$. We compute their relative distance and size as:
$$\left( \frac{2(x_j - x_i)}{h_i + h_j}, \frac{2(y_j - y_i)}{h_i + h_j}, \log\frac{h_i}{h_j}, \log\frac{w_i}{w_j}  \right)$$
We then concatenate this coordinate-based feature vector with the time difference  $t_j - t_i$ and relative appearance $\lVert\mathcal{N}_v^{enc}(a_j) - \mathcal{N}_v^{enc}(a_i) \rVert_2$ and feed it to a neural network $\mathcal{N}_e^{enc}$ in order to obtain the initial edge embedding $h_{(i, j)}^{(0)}$.
\subsection{Training and Inference} \label{inference}
\noindent{\bf Training Loss}.
To classify edges, we use an MLP with a sigmoid-valued single output unit, sharing weights with $\mathcal{N}_{e}^{w}$. 
For training, we use the weighted binary cross-entropy of our predictions over the embeddings produced in the last $m$ message passing steps, with respect to the target flow variables ($\mathcal{L}_t$). Since our classification problem is severely imbalanced, we weight positive terms in the loss with the inverse proportion of positive samples.

Masks are obtained by feeding the updated node features to a CNN that classifies each pixel of the RoI as foreground/background with a per-pixel sigmoid.   We define $\mathcal{L}_s$ as the average binary cross-entropy loss over the pixels. Similar to $\mathcal{L}_t$,  we compute this loss for the last $m$ message passing steps. Finally, we define a multi-task loss as $\mathcal{L} = \mathcal{L}_t + \mathcal{L}_s$, which is the main objective of our training.

\noindent{\bf Tracking Inference.}
During inference, we interpret the set of output values obtained from our model at the last message passing step as the solution to our data association problem. An easy way to obtain hard 0 or 1 decisions is to binarize the
output by thresholding. However, this procedure does not
generally guarantee that the flow conservation constraints in Equations~\ref{flow_in_constr} and \ref{flow_out_constr} are preserved.
In practice, thanks to the proposed time-aware update step, our method will satisfy over $98\%$ of the constraints on average when thresholding at 0.5. After that, a simple rounding scheme suffices to obtain a feasible binary output. We consider the significantly smaller subgraph consisting of nodes and edges involved in a violation of Equations~\ref{flow_in_constr} and \ref{flow_out_constr} and solve a linear program with it to ensure the satisfaction of constraints \ref{flow_in_constr} and \ref{flow_out_constr}. Given the high constraint satisfaction rate obtained by our model, the runtime of this procedure adds an insignificant computational overhead. As shown in \cite{mpntrack}, the linear program can be replaced by a simple heuristic rounding procedure without loss of tracking performance.

\noindent{\bf Mask prediction.} To obtain our final masks, we use a 7-layer CNN, denoted as $\tilde{\mathcal{N}}_{v}^{mask}$, to predict binary masks from the node mask features for each RoI. Our segmentation network $\tilde{\mathcal{N}}_{v}^{mask}$ receives two sets of features, namely the updated node features $\tilde{h}_i^{(l)}$ via attentive message passing, and the raw node features $\tilde{h}_i^{(0)}$ prior to message passing updates, which can be thought of as a skip connection:

\begin{align} \label{mask_net}
&    &mask_{i} &=  \tilde{\mathcal{N}}_{v}^{mask}([\tilde{h}_{(i)}^{(l)}, \tilde{h}_{(i)}^{(0)}])
\end{align}

\section{Experiments} \label{experiments_section}

In this section, we start by presenting ablation studies to understand the behavior of our model better. We then compare our model to the published methods on several datasets and show state-of-the-art results. For the MOT experiments, we use our default time-aware message passing framework, as introduced in our conference paper, and denote it as MPNTrack. For MOTS experiments, we extend our model to include also attentive message passing as explained in Section \ref{att_message_passing} and denote our model as \MPNSeg. Implementation details of the models are reported in Section \ref{imp_details}.

\subsection{\bf Evaluation Metrics.}

To evaluate our method, we report the CLEAR MOT \cite{clear}, IDF1 \cite{ristanieccvw2016} and HOTA \cite{luiten2020IJCV}, together with the percentage of mostly tracked (MT) and mostly lost targets (ML). For the CLEAR MOT metrics, we use the multiple object tracking accuracy (MOTA), which combines false positive detections (FP), false negative detections (FN), and identity switches (IDs) into a single score. Despite its widespread use, MOTA  accounts mostly for the quality of detections and is not affected significantly by identity preservation errors \cite{ristanieccvw2016}. IDF1, instead, is based on matching predicted trajectories to ground truth trajectories, instead of boxes, and therefore provides a better measure of data association performance. Lastly, the recently proposed HOTA score finds a balance between detection and data association performance by being decomposable into Detection Accuracy (DetA) and Association Accuracy (AssA). Hence, it can provide more clarity into the sources of errors committed by different trackers. 
\cite{voigtlaender2019mots} adapts CLEAR MOT metrics for MOTS by accounting for the segmentation masks. Specifically, they replace the bounding box IoU with a mask-based IoU and propose multi-object tracking and segmentation accuracy (MOTSA) and soft multi-object tracking and segmentation accuracy (sMOTSA). MOTSA utilizes the number of true positives that reaches an IoU of 0.5 whereas, sMOTSA accumulates the soft number of true positives to incorporate segmentation quality even more into MOTA. 

\subsection{\bf Datasets.}
\noindent{\bf MOTChallenge.} The multiple object tracking benchmark MOTChallenge
consists of several challenging pedestrian tracking sequences, with frequent occlusions and crowded scenes. 
The challenge includes four datasets {\it 2D MOT 2015}~\cite{lealarxiv2015}, {\it MOT16}~\cite{milanarxiv2016}, {\it MOT17}~\cite{milanarxiv2016} and {\it MOT20}~\cite{mot20}. They contain sequences lasting from 3 seconds to over 2 minutes, and with varying viewing angles, size, number of objects, camera motion, and frame rate. MOT20 is notable for its extreme crowdedness, with close to 150 pedestrians per frame, on average. MOTA and IDF1 scores are the most important metrics in this benchmark.

\noindent{\bf KITTI.} The KITTI Vision Benchmark Suite \cite{6248074} focuses on robotics applications and includes sequences of autonomous driving scenarios captured both in rural and urban areas and on highways. KITTI accommodates challenging computer vision benchmarks, including optical flow, visual odometry, and multi-object tracking. The tracking benchmark contains 21 sequences for training and 29 for testing, recorded at 10 frames per second. Not all sequences contain pedestrians, and those that do are characterized by their low object density. KITTI has recently incorporated HOTA as the main metric, and it also makes use of MOTA.

\noindent{\bf MOTS20 and KITTI-MOTS}~\cite{voigtlaender2019mots}. Recently, \cite{voigtlaender2019mots} extended both MOTChallenge and KITTI sequences for MOTS. MOTS20 consists of 4 training and 4 testing MOT16 sequences annotated with high-resolution instance masks. KITTI-MOTS adds segmentation masks for both pedestrians and cars to all KITTI sequences. As for metrics, sMOTSA and IDF1 are the main metrics to evaluate performance in MOTS20, whereas KITTI-MOTS bases its evaluation on domain-adapted HOTA that accounts for the mask quality. 

\noindent{\bf Human in Events (HiEve)}~\cite{hie}. The HiE dataset is a recently proposed benchmark focused on pedestrian tracking, detection, pose estimation, and action recognition in diverse surveillance scenarios, often characterized by heavy occlusions. It is the largest dataset currently available for pedestrian tracking, containing over 1.3 million annotated boxes across 19 sequences for training and 13 for testing, with an average trajectory length of over 480 frames. It uses two of the same metrics used in MOTChallenge, MOTA and IDF1, and does not accommodate HOTA.

\subsection{Implementation Details} \label{imp_details}

\noindent{\bf Network Models.} For the network $\mathcal{N}_v^{enc}$ used to encode detections appearances (see section  \ref{feature_encoding}), we employ a ResNet50\cite{He2016DeepRL} architecture pretrained on ImageNet \cite{imagenet_cvpr09}, followed by global average pooling and two fully-connected layers to obtain embeddings of dimension 256.
We train the network for the task of ReIdentification (ReID) jointly on three publicly available datasets: Market1501\cite{market_dataset}, CUHK03\cite{cuhk03_dataset} and DukeMTMC\cite{ristanieccvw2016}. Note that using external ReID datasets is a common practice among MOT methods \cite{TangAAS17,Kim_2018_ECCV, maACCV2019}.

Once trained, three new fully connected layers are added after the convolutional layers to reduce the embedding size of $\mathcal{N}_v^{enc}$ to 32.  To obtain secondary node features, we use a COCO pretrained ResNet50-FPN backbone \cite{He2016DeepRL, Lin_2017_CVPR}, referred as $\tilde{\mathcal{N}}_v^{enc}$. 

\noindent{\bf Data Augmentation.} To train our network, we sample batches of graphs. Each graph corresponds to small clips with a fixed number of frames. For MOT experiments, we use 15 frames per graph sampled at six frames per second for static sequences and nine frames per second for those with a moving camera. For MOTS experiments, on the other hand, we use 30 frames per graph without any sampling. We perform data augmentation by randomly removing nodes from the graph, hence simulating missed detections, and randomly shifting bounding boxes.

 \noindent{\bf Training.} We have empirically observed that additional training of the ResNet blocks provides no significant increase in performance, but carries a significantly larger computational overhead. Hence, during training, we freeze all convolutional layers and train jointly all of the remaining model components.
 We train for 15000 iterations for MOT and fifteen epochs for MOTS with a learning rate of $3 \cdot 10^{-4}$, weight decay term of $10^{-4}$ and an Adam Optimizer with $\beta_1$ and $\beta_2$ set to $0.9$ and $0.999$, respectively. 
 
 \noindent{\bf Batch Processing.} We process videos offline in batches of $n$ frames, with $n-1$ of those overlapping, to ensure that the maximum time distance between two connected nodes in the graph remains stable across the whole graph. We prune graphs by connecting two nodes only if both are among the top-$K$ mutual nearest neighbors according to the ResNet features. We set $n=15$, $K=50$ for MOT and $n=30$, $K=150$ for MOTS experiments.
Each batch is solved independently by our network, and for overlapping edges and masks between batches, we average the predictions coming from all graph solutions. To fill gaps in our trajectories, we perform simple bilinear interpolation over the missing frames.

\noindent{\bf Baseline.}
Recently, \cite{tracktor} have shown the potential of detectors for simple data association, establishing a new baseline for MOT. We exploit this in our MOT version and preprocess all sequences by first running \cite{tracktor} on public detections, which allows us to be  fully comparable to all methods on MOTChallenge. Note that this procedure has been adopted by several methods in the recent literature \cite{lpcmot, pmlr-v119-hornakova20a, graph_matching}. One key drawback of \cite{tracktor} is its inability to fill in gaps, hence failing in properly recovering identities through occlusions. As we will show, this is exactly where our method excels. For the MOTS domain, however, we don't adopt this scheme with \cite{tracktor} and we use our model independently. 

\noindent{\bf Detections} For all MOT benchmarks in MOTChallenge and HiEve, we use the provided detections to ensure a fair comparison with other methods. For MOTS20 and KITTI benchmarks, we use bounding box detections obtained with a Mask R-CNN~\cite{He_2017_ICCV} with ResNeXt-152 backbone \cite{Xie_2017_CVPR} trained on ImageNet and COCO \cite{10.1007/978-3-319-10602-1_48}, as these benchmarks do not accommodate public detections. 

\begin{figure}
\includegraphics[width=0.50\textwidth]{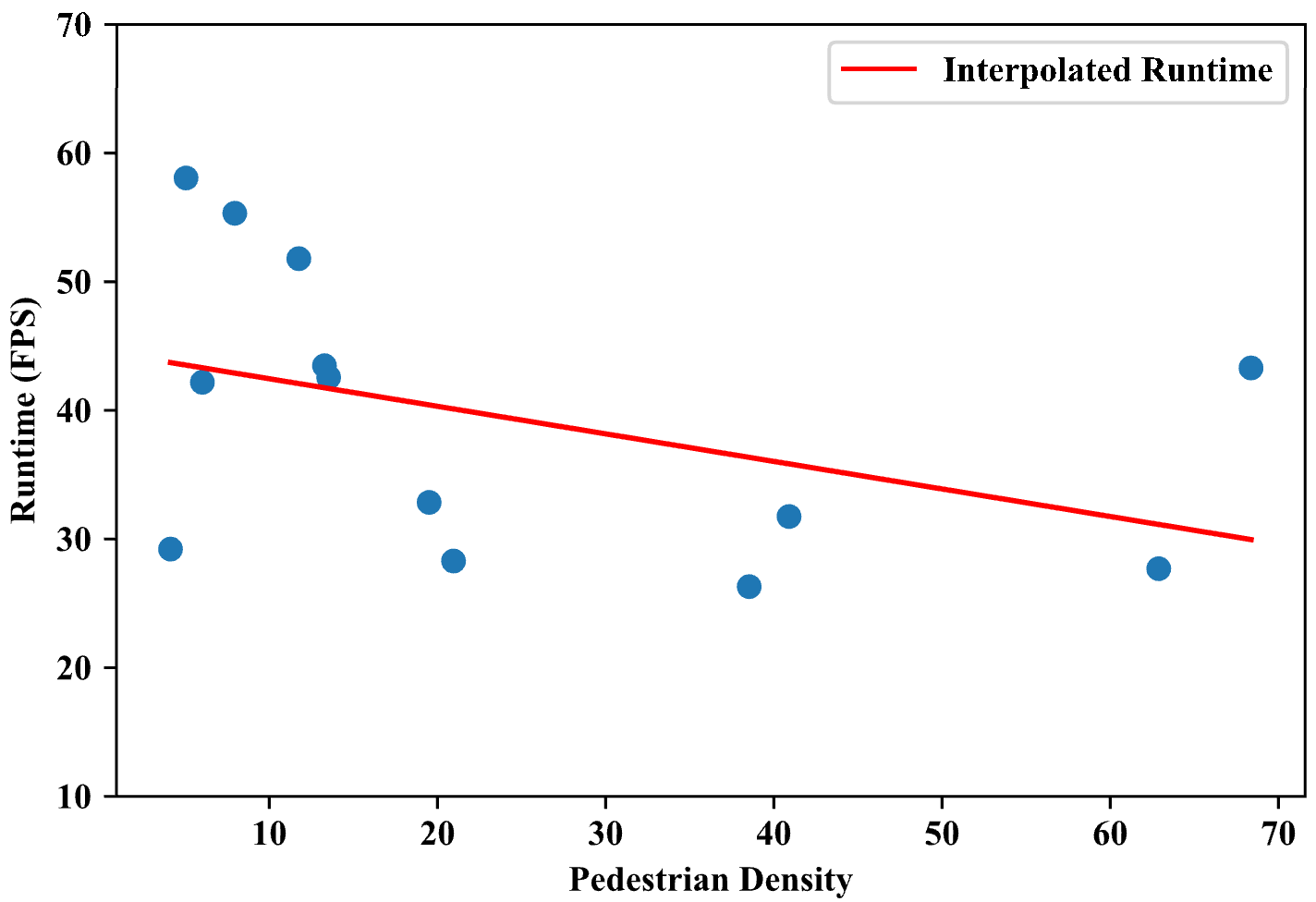}
\centering
\caption{We report our method's runtime in frames-per-second over sequences with varying pedestrian densities, as measured by average detections per frame.} \label{seq_runtime}
\end{figure}

\noindent{\bf Runtime.} We build our graph on the output of \cite{tracktor} for MOT. Hence, we take also its runtime into account. Our method, on its own, runs at 35fps, while \cite{tracktor} without the added re-ID head runs at 8fps, which gives the reported average of 6.5fps on a single Nvidia P5000 GPU, running on a machine with 8 3.6GHz CPU cores. When we incorporate the attentive message passing scheme and segmentation head, our unified framework runs at 2.3fps on MOTS20. To provide further analysis, in Figure \ref{seq_runtime} we report our method's runtime in fps over MOT15, MOT16, and MOT20 test sequences depending on their pedestrian density (in average detections per frame). While our method is naturally slower in denser sequences, it still shows a very competitive runtime in very crowded scenes containing over 60 pedestrians per frame. 

\subsection{Ablation Study} \label{ablation_study}
In this section, we aim to answer five main questions towards understanding our model. Firstly, we compare the performance of our time-aware neural message passing updates with respect to the time-agnostic vanilla node update described in Section \ref{vanilla_mpns}.
Secondly, we assess the impact of the number of message passing steps in network training to the overall tracking performance.
Thirdly, we investigate how different information sources, namely, appearance embeddings from our CNN and relative position information, affect different evaluation metrics. Then, we quantify the impact of the attentive message passing explained in Section \ref{att_message_passing} by exploring the effect of the features used for mask prediction. Finally, we compare our tracking-only model and unified model to demonstrate the effect of jointly training for tracking and segmentation.

\noindent{\bf Experimental Setup}. We conduct all of our experiments with the training sequences of the MOT15, MOT17, and MOTS20 datasets. To evaluate our models on MOT datasets, we split MOT17 sequences into three sets (sequences 2, 10 and 3; 4 and 11; and 5 and 9), and use these to test our models with 3-fold cross-validation. We then report the best overall MOT17 metrics obtained during validation. For MOTS results, we perform 4-fold cross-validation on MOTS20 and report the mean scores obtained in the best epochs by leaving one sequence out at a time.

\noindent{\bf Time-Aware Message Passing}. We investigate how our proposed time-aware node update affects performance. For a fair comparison, we also add the skip connection with respect to initial edge features to the vanilla update, and increase the parameter count of its node update MLP to match the overall parameter count of the time-aware model. Still, we observe a significant improvement in almost all metrics, including close to 3 points in IDF1. 
As we expected, our model is particularly powerful at linking detections, since it exploits neighboring information and graph structure, making the decisions more robust, and hence producing significantly fewer identity switches. 
We also report the percentage of constraints that are satisfied when directly thresholding our model's output values at 0.5. Remarkably, our method with time-aware node updates is able to produce almost completely feasible results automatically, i.e., 98.8\% constraint satisfaction, while the baseline has only 83.2\% satisfaction. This demonstrates its ability to capture the MOT problem structure.
%!TEX root = ../1404.tex

\begin{table}
\center
\tabcolsep=0.11cm

    \resizebox{\columnwidth}{!}{
    \begin{tabular}{l c  c c c c c c c c}
     \toprule
     Arch. & MOTA $\uparrow$ & IDF1 $\uparrow$ & MT $\uparrow$ & ML $\downarrow$ & FP $\downarrow$ & FN $\downarrow$ & ID Sw. $\downarrow$ &Constr. $\uparrow$\\ [0.5ex] 
     \midrule
     
                   %MOTA   IDF1    MT     ML     FP       FN     IDs
     Vanilla  & 63.2  & 67.1 & 586 &  372 & 3239 & 119853 & 917 & 83.2 \\
     T. aware  & 64.0 & 70.0 & 648 &  362 & 6169 & 114509 & 602   & 98.8 \\

     \midrule

    \end{tabular}}

\caption{We investigate how our proposed update improves tracking performance with respect to a vanilla MPN. \textit{Vanilla} stands for a basic MPN, \textit{T. aware} denotes our proposed time-aware update. The metric \textit{Constr} refers to the percentage of flow conservation constraints satisfied on average over entire validation sequences.}
\vspace{-0.2cm}
\label{tab:ablation_arch}

\end{table} \label{ablation_arch}

\begin{figure}[h]
        \centering
           \subfloat[MPNTrack]{%
              \includegraphics[width=0.50\textwidth]{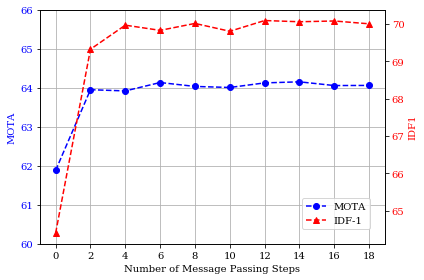}
              \label{num_mp_steps}
           } 
           \vspace{0cm}
           \subfloat[\MPNSeg]{%
              \includegraphics[width=0.50\textwidth]{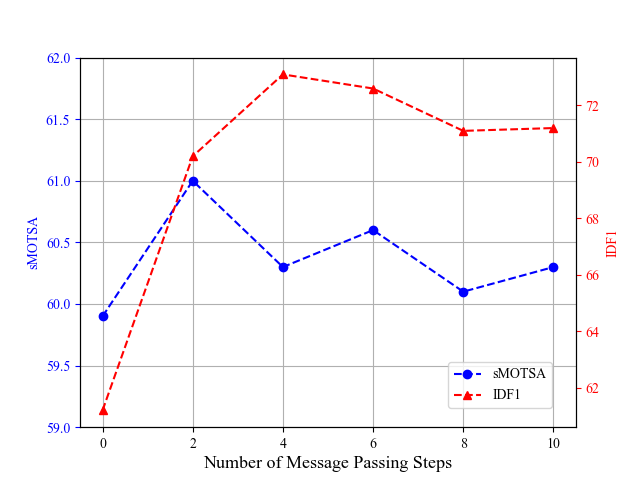}
              \label{mots_steps}
           }
           \caption{We report the evolution of IDF1, MOTA and sMOTSA when training networks with an increasing number of message passing steps for MPNTrack and \MPNSeg.}
           \label{3}
\end{figure}

\noindent{\bf Number of Message Passing Steps}. 
Intuitively, increasing the number of message passing steps $L$ allows each node and edge embedding to encode further context and gives edge predictions the ability to be iteratively refined.
Hence, one would expect $L$ values greater than zero to yield better-performing networks. We test this hypothesis in Figure \ref{num_mp_steps} by training networks with a fixed number of message passing steps, from 0 to 18. We use the case $L=0$ as a baseline in which we train a binary classifier on top of our initial edge embeddings, and hence, no contextual information is used.
As expected, we see a clear upward tendency for both IDF-1 and MOTA. Moreover, we observe a steep increase in both metrics from zero to two message passing steps, which demonstrates that the most improvement is obtained when switching from pairwise to high-order features in the graph. 
 We also note that the upwards tendency stagnates after four message passing steps and shows no improvement after twelve message passing steps. We repeat the same experiment for \MPNSeg and report our findings in Figure~\ref{mots_steps}. Similar to our previous findings, we observe an increase in IDF1 and sMOTSA from zero to two message passing steps. The upward trend continues until four steps, and the performance stagnates between four to six steps. Hence, we use $L=12$ in our final configuration for MOT and $L=4$ for MOTS.

\noindent{\bf Effect of the Features}. In the MOT setting, our model receives two main streams of information: (i) appearance information from a CNN and (ii) geometry features from an MLP encoding relative position between detections.
We test their usefulness by experimenting with combinations of three groups of features for edges: time difference, relative position, and the euclidean distance in CNN embeddings between the two bounding boxes. Results are summarized in Table \ref{tab:ablation_features_table}.
We highlight the fact that relative position seems to be a key component of overall performance since its addition yields the largest relative performance increase.
Nevertheless, CNN features are powerful to reduce the number of false positives and identity switches; hence, we use them in our final configuration. 
%!TEX root = ../1404.tex

\begin{table}
\center
\tabcolsep=0.11cm

    \resizebox{\columnwidth}{!}{
    \begin{tabular}{l c  c c c c c c c c}
     \toprule
     Edge Feats. & MOTA $\uparrow$ & IDF1 $\uparrow$ & MT $\uparrow$ & ML $\downarrow$ & FP $\downarrow$ & FN $\downarrow$ & ID Sw. $\downarrow$ \\ [0.5ex] 
     \midrule

     Time & 58.8 &  52.6 &  529 &  372 &  13127 &  122800 &  2962 \\

     Time+CNN & 62.3 &  64.5 &  641 &  363 &   10923 &  115375 &   812 \\
     
     Time+Pos & 63.6 &  68.7 &  631 &  365 &   6308 &  115506 &   895 \\
    
     Time+Pos+CNN  & 64.0 & 70.0 & 648 &  362 & 6169 & 114509 & 602    \\

     \midrule

    \end{tabular}}

\caption{We explore combinations of three sources of information for edge features: time difference in seconds (Time), relative position features (Pos) and the Euclidean distance between CNN embeddings of the two detections (CNN).}
\vspace{-0.2cm}
\label{tab:ablation_features_table}
\end{table}
 \label{ablation_features_table}

% \begin{figure} 

% \includegraphics[width=0.50\textwidth]{final_num_mp_steps.png}
% \centering
% \caption{We report the evolution of IDF-1 and MOTA when training networks with an increasing number of message passing steps for MPNTrack.} \label{num_mp_steps}
% \end{figure}

% \begin{figure} 
% \includegraphics[width=0.50\textwidth]{mots_steps_v2.png}
% \centering
% \caption{We report the evolution of IDF-1 and sMOTSA when training networks with an increasing number of message passing steps for MPNSeg.} \label{mots_steps}
% \end{figure}

\noindent{\bf Attentive Message Passing}. We now evaluate the effect of our proposed attentive message passing updates for multi-object tracking and segmentation. 
 As can be seen in Equation \ref{mask_net}, our model receives both updated node features from message passing and raw node features via a skip connection for mask prediction. 
 In Table  \ref{tab:ablation_mots_features_table}, we experiment with the combinations of these features. 
 In addition, we compare these results with an additive message passing scheme that simply aggregates features from neighbors via summation instead of using attention. Note that only using the raw features via skip connections is equivalent to eliminating the message passing on mask features from our model and corresponds to using an independent segmentation network on top of our tracking-only version (MPNTrack), which serves as a baseline. Our graph-based framework already achieves competitive results while using skip connections with raw features alone. Performing message passing updates by simply summing neighboring node features is commonly used as an aggregation function in MPNs, however, this approach can not capture a rich visual representation required for our setting, as indicated by the lower sMOTSA score. We speculate that this drop is caused by the fact that incorporating segmentation features from neighboring nodes with equal weights harms our mask quality due to the outweighing caused by the large number of neighboring nodes belonging to different trajectories. With attention-based updates, we improve upon additive updates, highlighting the importance of taking identity information into account while performing message passing updates. Combining both raw features and attentive updates further boosts our performance, as indicated by the 1.1 points increase in IDF1 score and 15\% drop in ML. Overall, while identity preservation scores get a significant boost from our attentive message passing module, we note that mask-quality related scores i.e., sMOTSA stay comparable to those of our baseline. We conclude that the main advantage of our joint framework lies in its ability to boost association performance by exploiting the rich appearance information that segmentation features provide.

%!TEX root = ../1404.tex

\begin{table}
\center
\tabcolsep=0.08cm

    \resizebox{\columnwidth}{!}{
    \begin{tabular}{l c c c c c c c c c c c c}
     \toprule
     Method & Raw & Upd. & sMOTSA$\uparrow$ & IDF1$\uparrow$ & MOTSA$\uparrow$  & MT$\uparrow$ & ML$\downarrow$ & ID Sw.$\downarrow$ & MOTSP$\uparrow$ & TP$\uparrow$ \\ [0.5ex] 
     \midrule

     MPNTrack + Seg & \cmark & & 60.5 & 71.8 &  75.1 & 130 & 20 & 252 & 82.1 & 21864 \\
     
    %   \MPNSeg-Add &  & additive & 57.3 & 72.2 & 73.2 & 127 & 21 & 227 & 80.2 & 21651 \\

    %   \MPNSeg-Att & & attention & 60.7 &  72.0 &  75.1 &  130 &  20 &  228 & 82.3 & 21903 \\
     
    %   \MPNSeg-Att+R & \cmark & attention & 60.3 &  73.1 &  74.9 &  128 &  17 &  223 & 82.0 & 21858 \\
     
    \MPNSeg-Add &  & Add. & 57.3 & 72.2 & 73.2 & 127 & 21 & 227 & 80.2 & 21651 \\

     \MPNSeg-Att & & Att. & 60.7 &  72.0 &  75.1 &  130 &  20 &  228 & 82.3 & 21903 \\
     
     \MPNSeg-Att+R & \cmark & Att. & 60.3 &  73.1 &  74.9 &  128 &  17 &  223 & 82.0 & 21858 \\
    
    %  \MPNSeg &  & Add. & 57.3 & 72.2 & 73.2 & 127 & 21 & 227 & 80.2 & 21651 \\

    %  \MPNSeg & & Att. & 60.7 &  72.0 &  75.1 &  130 &  20 &  228 & 82.3 & 21903 \\
     
    %  \MPNSeg & \cmark & Att. & 60.3 &  73.1 &  74.9 &  128 &  17 &  223 & 82.0 & 21858 \\

     \midrule

    \end{tabular}}

\caption{We examine the contributions of the features and update schemes used for segmentation: initial node features before message passing updates (Raw) and updated node features (Upd.) after additive (Add.) and attentive (Att.) message passing steps.}
\vspace{-0.2cm}
\label{tab:ablation_mots_features_table}
\end{table}
 \label{ablation_mots_features_table}

\noindent{\bf Joint Training for Tracking and Segmentation}. 
We compare MPNTrack and \MPNSeg in terms of data association performance to demonstrate the effect of joint training. For a fair comparison, we use the same parameters for both models and train them on MOTS20. Based on a shared set of ground truth, we use bounding box-based annotations for MPNTrack and mask-based annotations for \MPNSeg. After training, we disable the mask-related components of \MPNSeg. We base the evaluation on bounding boxes and report MOT metrics. As shown in Table~\ref{tab:ablation_models}, we observe +0.4 IDF1 and +0.2 MOTA improvements with our unified model. These observations suggest that association performance benefits from mask information and joint training within our unified framework.

%!TEX root = ../1404.tex

\begin{table}
\center
\tabcolsep=0.11cm

    \resizebox{\columnwidth}{!}{
    \begin{tabular}{l c  c c c c c c c c}
     \toprule
     Edge Feats. & MOTA $\uparrow$ & IDF1 $\uparrow$ & MT $\uparrow$ & ML $\downarrow$ & FP $\downarrow$ & FN $\downarrow$ & ID Sw. $\downarrow$ \\ [0.5ex] 
     \midrule

     MPNTrack & 62.1 &  68.2 &  134 &  17 &   5106 &  4867 &   232 \\

     \MPNSeg & 62.3 &  68.6 &  129 &  20 &   4729 &  5143 &  271 \\

     \midrule

    \end{tabular}}

\caption{We compare our tracking-only model (MPNTrack) and unified model (\MPNSeg) to demonstrate the effect of jointly training for tracking and segmentation. We perform bounding-box-based evaluation and report MOT metrics.}
\vspace{-0.2cm}
\label{tab:ablation_models}
\end{table}
 \label{ablation_models}

\subsection{Benchmark Evaluation}

\subsubsection{Multi-Object Tracking}
%!TEX root = ../1404.tex

\begin{table*}
\center
\tabcolsep=0.11cm

    \resizebox{\columnwidth*2}{!}{
    \begin{tabular}{l c c c c c c c c c c c }
     \toprule
     Method & On/Off& MOTA $\uparrow$ & IDF1 $\uparrow$ & HOTA $\uparrow$ & MT $\uparrow$ & ML $\downarrow$ & FP $\downarrow$ & FN $\downarrow$ & ID Sw. $\downarrow$  & Hz  $\uparrow$\\ [0.5ex] 
     \midrule
     \multicolumn{9}{c}{2D MOT 2015~\cite{lealarxiv2015}    } \\
     \midrule
     JointMC~\cite{8493320} & Offline & 35.6 & 45.1& 34.6 & 23.2 & 39.3 & 10580 & 28508 & 457 & 0.6\\
     AMIR15~\cite{Sadeghian_2017_ICCV}& Online & 37.6 & 46.0& 33.2 & 15.8 & 26.8 & 7933 & 29397 & 1026 & 1.9 \\
     
      STRN~\cite{spatio_temporalrelation_networks}& Online & 38.1 & 46.6& 33.5 & 11.5 & 33.4 & 5451 & 31571 & 1033  & 13.8\\
      DeepMOT\textsuperscript{$\dagger$}~\cite{Xu_2020_CVPR}& Online & 44.1 &  46.0 & 36.3 & 17.2 &  26.6  & 6085 & 26917 & 1347 & 1.6\\
      Tracktor++v2\textsuperscript{$\dagger$}~\cite{tracktor}& Online & 46.6 &  47.6 & 37.6 & 18.2 &  27.9  & 4624 & 26896 & 1290 & 1.8\\ 

      \textcolor{gray}{Lif\_T\textsuperscript{$\dagger$}~\cite{pmlr-v119-hornakova20a}} & \postcvpr{Offline} &  \postcvpr{\textbf{52.5}} &  \postcvpr{\textbf{60.0}} & \postcvpr{\textbf{46.0}} & \postcvpr{\textbf{33.8}} & \postcvpr{\textbf{25.8}}  & \postcvpr{\textbf{6837}} & \postcvpr{\textbf{21610}} & \postcvpr{730} & \postcvpr{1.5} \\
      MPNTrack\textsuperscript{$\dagger$} (Ours) & Offline &  \textbf{51.5} &  \textbf{58.6} & \textbf{45.0} & \textbf{31.2} &  \textbf{25.9}  & \textbf{7620} & \textbf{21780} & \textbf{375} & 6.5 \\
     \midrule
     \multicolumn{9}{c}{MOT16~\cite{milanarxiv2016}} \\
     \midrule

     FWT~\cite{HenschelLCR17} &Offline &47.8 & 44.3 & 35.7 & 19.1 & 38.2 & 8886 & 85487 & 852 & 0.6 \\
     STRN~\cite{spatio_temporalrelation_networks}&Online & 48.5 & 53.9 & 39.7 & 17.0 & 34.9 & 9038 & 84178 & 747  & 13.5\\
     LMP~\cite{TangAAS17} &Offline &48.8 & 51.3 & 41.0 & 18.2 & 40.1 & 6654 & 86245 & 481  & 0.5\\

     \postcvpr{BLSTM\_MTP\_O~\cite{kim2021cvpr}} & \postcvpr{Online} & \postcvpr{48.3} & \postcvpr{53.5} & \postcvpr{39.7} & \postcvpr{17.0} & \postcvpr{38.7} & \postcvpr{9792} & \postcvpr{83707} & \postcvpr{735}  & \postcvpr{21.0}\\
     Tracktor++v2\textsuperscript{$\dagger$} \cite{tracktor}& Online & 56.2 &  54.9& 44.8  & 20.7 & 35.8 &  \textbf{2394} & 76844 & 617 & 1.8\\
     \postcvpr{LPC\_MOT\textsuperscript{$\dagger$}~\cite{lpcmot}}& \postcvpr{Offline} & \postcvpr{58.8} &  \postcvpr{\textbf{67.6}} & \postcvpr{51.7} & \postcvpr{\textbf{27.3}} & \postcvpr{35.0} &  \postcvpr{6167} & \postcvpr{68432} & \postcvpr{435} & \postcvpr{4.3}\\
     \postcvpr{Lif\_T\textsuperscript{$\dagger$}~\cite{pmlr-v119-hornakova20a}}& \postcvpr{Offline} & \postcvpr{61.3} &  \postcvpr{64.7} & \postcvpr{\textbf{50.8}} & \postcvpr{27.0} & \postcvpr{34.0} &  \postcvpr{4844} & \postcvpr{65401} & \postcvpr{389} & \postcvpr{0.5}\\
     \postcvpr{TMOH\textsuperscript{$\dagger$}~\cite{tmoh}}& \postcvpr{Online} & \postcvpr{\textbf{63.2}} &  \postcvpr{63.5} & \postcvpr{50.7} & \postcvpr{27.0} & \postcvpr{\textbf{31.0}} &  \postcvpr{3122} & \postcvpr{\textbf{63376}} & \postcvpr{635} & \postcvpr{0.7}\\
     MPNTrack\textsuperscript{$\dagger$} (Ours)& \postcvpr{Offline} & \textbf{58.6} & \textbf{61.7} & \textbf{48.9} & \textbf{27.3} &  \textbf{34.0} & 4949 & \textbf{70252} &  \textbf{354} & 6.5 \\
     \midrule
     \multicolumn{9}{c}{ MOT17~\cite{milanarxiv2016}   } \\
     \midrule

     jCC~\cite{8493320} & Offline & 51.2 & 54.5 & 42.5 & 20.9 & 37.0 & 25937 & 247822 & 1802 &1.8\\
     FAMNet \cite{famnet}&Online &52.0 & 48.7 & -- & 19.1 & 33.4 & 14138 & 253616 & 3072 & --\\     
     JBNOT \cite{JBNOT}&Offline &  52.6 & 50.8& 41.3 & 19.7 & 35.8 & 31572 & 232659 & 3050 & 5.4 \\
     Tracktor++v2\textsuperscript{$\dagger$}~\cite{tracktor}&Online& 56.3 & 55.1& 44.8 & 21.1 & 35.3 & \textbf{8866} & 235449 & 1987 & 1.8\\
     \postcvpr{LPC\_MOT\textsuperscript{$\dagger$}~\cite{lpcmot}} & \postcvpr{Offline}& \postcvpr{59.0} & \postcvpr{\textbf{66.8}} & \postcvpr{\textbf{51.5}} & \postcvpr{\textbf{29.9}} & \postcvpr{33.9} & \postcvpr{23102} & \postcvpr{206948} & \postcvpr{\textbf{1122}} & \postcvpr{4.8} \\
     \postcvpr{Lif\_T\textsuperscript{$\dagger$}~\cite{pmlr-v119-hornakova20a}} & \postcvpr{Offline}& \postcvpr{60.5} & \postcvpr{65.6} & \postcvpr{51.3} & \postcvpr{27.0} & \postcvpr{33.6} & \postcvpr{14966} & \postcvpr{206619} & \postcvpr{1189} & \postcvpr{0.5} \\
     \postcvpr{CenterTrack~\cite{10.1007/978-3-030-58548-8_28}} & \postcvpr{Online} & \postcvpr{61.5} & \postcvpr{59.6} & \postcvpr{48.2} & \postcvpr{26.4} & \postcvpr{31.9} & \postcvpr{14076} & \postcvpr{\textbf{200672}} & \postcvpr{2583} & \postcvpr{17.0} \\
     \postcvpr{TMOH\textsuperscript{$\dagger$}~\cite{tmoh}}& \postcvpr{Online} & \postcvpr{\textbf{62.1}} & \postcvpr{62.8} & \postcvpr{50.4} & \postcvpr{26.9} & \postcvpr{\textbf{31.4}} & \postcvpr{10951} & \postcvpr{201195} & \postcvpr{1897} & \postcvpr{0.7} \\
     MPNTrack\textsuperscript{$\dagger$} (Ours)& Offline & \textbf{58.8} & \textbf{61.7} & \textbf{49.0} & \textbf{28.8} & \textbf{33.5} & 17413 & \textbf{213594} & \textbf{1185} & 6.5 \\
     \midrule
     \multicolumn{9}{c}{ MOT20~\cite{mot20}   } \\

     \midrule
     SORT ~\cite{sort}&Online  &42.7 & 45.1 & 36.1 & 16.7 & 26.2 & 27521 & 264694 & 4470 & 57.3\\ 
     Tracktor++v2\textsuperscript{$\dagger$} ~\cite{tracktor}&Online   &52.6 & 52.7 & 42.1 & 29.4 & 26.7 & \textbf{6930} & 236680 & 1548 & 1.8\\
     \postcvpr{LPC\_MOT\textsuperscript{$\dagger$}~\cite{lpcmot}}&\postcvpr{Offline} &\postcvpr{56.3} & \postcvpr{\textbf{62.5}}& \postcvpr{\textbf{49.0}} & \postcvpr{34.1} & \postcvpr{25.2} & \postcvpr{11726} & \postcvpr{213056} & \postcvpr{1562} & \postcvpr{0.7}\\     

     \postcvpr{TMOH\textsuperscript{$\dagger$}~\cite{tmoh}}&\postcvpr{Online} & \postcvpr{\textbf{60.1}} & \postcvpr{61.2}& \postcvpr{48.9} & \postcvpr{\textbf{46.7}} & \postcvpr{\textbf{15.2}} & \postcvpr{38043} & \postcvpr{\textbf{165899}} & \postcvpr{2342} & \postcvpr{0.6} \\
     MPNTrack\textsuperscript{$\dagger$} (Ours)& Offline&\textbf{57.6} & \textbf{59.1} & \textbf{46.8} & \textbf{38.2} & \textbf{22.5} & 16953 & \textbf{201384} & \textbf{1210} & 6.5 \\
     
     \bottomrule
    \end{tabular}
}

\caption{Comparison of our method with state-of-the art on the \textbf{MOTChallenge test} sets. Methods written in grey were published after our CVPR2020 work. Methods with \textsuperscript{$\dagger$} after their name also used Tracktor-based preprocessing on their input boxes.}
\vspace{-0.2cm}
\label{tab:mot}

\end{table*} \label{mot_metrics_results}

%!TEX root = ../1404.tex

\begin{table*}
\center
\tabcolsep=0.11cm

    \resizebox{\columnwidth*2}{!}{
    \begin{tabular}{l c c c c c c c c c c}
     \toprule
     Method & HOTA $\uparrow$ & DetA $\uparrow$ & AssA $\uparrow$ & MOTA $\uparrow$ & MT (\%) $\uparrow$ & ML(\%) $\downarrow$ & FP $\downarrow$ & FN $\downarrow$ & ID Sw. $\downarrow$ \\ [0.5ex] 
     \midrule
     \multicolumn{10}{c}{2D Methods} \\
     \midrule
     
     NOMT \cite{choiiccv2015} & 36.3 &  31.9 & 41.6 & 36.5 &  19.2  &  43.6 & 12319 & \textbf{2249} &  \textbf{127} \\
     \postcvpr{CenterTrack~\cite{10.1007/978-3-030-58548-8_28}} & \postcvpr{40.4} &  \postcvpr{44.5} & \postcvpr{35.4} & \postcvpr{53.8} &  \postcvpr{21.3}  &  \postcvpr{34.7} & \postcvpr{8061} & \postcvpr{2201} &  \postcvpr{425} \\
     \postcvpr{Quasi-Dense~\cite{qdtrack}} & \postcvpr{41.1} &  \postcvpr{\textbf{44.8}} & \postcvpr{\textbf{38.1}}  & \postcvpr{55.6} &  \postcvpr{31.3}  &  \postcvpr{20.3} & \postcvpr{8493} & \postcvpr{\textbf{1309}} &  \postcvpr{487} \\
     %\midrule
     MPNTrack (Ours) & \textbf{45.3} &  \textbf{43.7} &  \textbf{47.3}  & \textbf{46.2}   &  \textbf{44.0}  & \textbf{10.3} &  \textbf{6956} &  5096 &  397 \\ %Thresh 0.6
    \midrule
     \multicolumn{10}{c}{3D Methods} \\
     \midrule
     
     CIWT \cite{Osep17ICRA} & 33.9 &  \textbf{34.0} & 34.1 & 42.1 &  14.1  &  \textbf{35.1} & 11821 & \textbf{1149} & 433 \\
     JRMOT \cite{jrmot} & 34.1 &  29.6 & \textbf{39.5} & \textbf{45.3} &  13.1 &  47.4 & \textbf{10207} & 1822 &  \textbf{179} \\
     AB3DMOT \cite{Weng2020_AB3DMOT} & \textbf{35.6} &  33.0 &  38.6  &  38.9  &  \textbf{17.2}  & 41.2 & 11744 &   2135 & 259 \\
     \postcvpr{EagerMOT~\cite{Kim21ICRA}} & \postcvpr{\textbf{39.4}} &  \postcvpr{\textbf{40.6}} &  \postcvpr{\textbf{38.7}} & \postcvpr{\textbf{49.8}}  &  \postcvpr{\textbf{27.5}}  & \postcvpr{\textbf{24.1}} &  \postcvpr{\textbf{8959}} & \postcvpr{2161} &  \postcvpr{496} \\

     \bottomrule
    \end{tabular}}

\caption{Comparison of our method with state-of-the-art on \textbf{KITTI-tracking test} set. Methods written in grey were published after our CVPR2020 work.
}
\vspace{-0.2cm}
\label{tab:kitti}

\end{table*}
%!TEX root = ../1404.tex

\begin{table*}
\center
\tabcolsep=0.11cm

    \resizebox{\columnwidth*2}{!}{
    \begin{tabular}{l c c c c c c c}
     \toprule
     Method & MOTA $\uparrow$ & IDF1 $\uparrow$ & MT(\%) $\uparrow$  & ML (\%) $\downarrow$ & FP$\downarrow$ & FN $\downarrow$ & ID Sw. $\downarrow$ \\ [0.5ex] 
     \midrule
     DeepSORT~\cite{deepsort} & 27.2 &  28.6 & 8.5  &  41.4  &  \textbf{5894} &  42668 & 2220\\
     \postcvpr{CenterTrack~\cite{10.1007/978-3-030-58548-8_28}}& \postcvpr{31.1} &  \postcvpr{41.8} & \postcvpr{8.6}  &  \postcvpr{27.9}  &  \postcvpr{10014} &  \postcvpr{35253} & \postcvpr{2767}\\
     \postcvpr{TPM~\cite{tpm}}  & \postcvpr{33.6} &  \postcvpr{37.7} & \postcvpr{10.7}  &  \postcvpr{31.2}  &  \postcvpr{6595} &  \postcvpr{35395} & \postcvpr{4287} \\

     \postcvpr{GMPHD-ReId~\cite{gmphd_reid}} & \postcvpr{31.3} &  \postcvpr{37.7} & \postcvpr{\textbf{36.0}}  &  \postcvpr{24.3}  &  \postcvpr{17309} &  \postcvpr{26158} & \postcvpr{4392}\\
     \postcvpr{STPP~\cite{stpp}} & \postcvpr{37.5} &  \postcvpr{40.2} & \postcvpr{20.4}  &  \postcvpr{29.8}  &  \postcvpr{7395} &  \postcvpr{31638} & \postcvpr{4536}\\
     \postcvpr{SiamMOT~\cite{shuai2021siammot}} & \postcvpr{\textbf{53.2}} &  \postcvpr{51.7} & \postcvpr{26.7}  &  \postcvpr{27.5} &  \postcvpr{\textbf{2837}} &  \postcvpr{28485} & \postcvpr{1308}\\
     %\midrule
     MPNTrack (Ours) & \textbf{48.0} &  \textbf{53.3} &  \textbf{33.6}  &  \textbf{23.8}  & 7756 &  \textbf{27236} &  \textbf{1243} \\ %Thresh 0.6

     \bottomrule
    \end{tabular}}

\caption{Comparison of our method with state-of-the-art on the \textbf{Human in Events test} dataset. Methods written in grey were published after our CVPR2020 work.
}
\vspace{-0.2cm}
\label{tab:hie}

\end{table*}

\noindent{\bf MOTChallenge}. In Table \ref{tab:mot}, we report our results in all four datasets in the MOTChallenge. We achieve competitive results and retain one of the fastest runtimes among published methods. We note that we are outperformed by two graph-based methods that were published after our approach: Lif\_T and LPC\_MOT. Lif\_T, also follows a graph-based formulation, however, it uses a complex optimization scheme and additional engineered features such as DeepMatching \cite{deepmatching}. As a result, it is over one order of magnitude slower than our approach. For LPC\_MOT, we note that it builds over our work, and it also combines a message passing network with a graph-based approach.  However, it replaces our network flow formulation with a significantly more involved multiple-hypothesis-based approach. We note that its runtime slows down significantly in crowded sequences such as those in MOT20, while our method retains the same speed and is, again, one order of magnitude faster.  Lastly, we also are outperformed by a regression-based method that was published after our approach: TMOH. It follows a bounding box regression-based approach with a stronger backbone and is, therefore, able to reduce the amount of false negative detections in a way that is not available to us. Note, however, that it is also significantly slower than our method. Moreover, it follows an orthogonal direction to our approach, and potentially, both methods could be combined, similarly to how we integrated Tracktor into our approach. 

\noindent{\bf KITTI}. In Table \ref{tab:kitti} we report our pedestrian tracking results among published methods using 2D inputs on KITTI (i.e. no Lidar nor depth), and observe that we surpass all previous methods in terms of HOTA by a significant margin (+4.2 points). Our improvement is due to the data association capabilities of our model, as can be seen by a +5.7 improvement with respect to the previous best association accuracy (AssA). We also note that our detection accuracy (DetA) is slightly lower than that of other newer methods, since we used off-the-shelf detections, and our method focuses solely on association. This can be further observed in our relatively lower MOTA, which is caused by a relatively larger number of False Negatives in the detections we used. However, as it can be seen from our overall HOTA score, our method has significantly superior overall tracking performance. These results show the versatility of our approach, and its ability to perform well in a setting for which it is not specialized: non-crowded autonomous driving scenes.

\noindent{\bf Human in Events}. In Table \ref{tab:hie} we report our results on the HiEve dataset. For a fair comparison, we exclude from this table competitors of the MM20' Grand Challenge \cite{hie}, as they used multiple additional datasets for training, as well as model ensembles. Among published methods, we observe that our model achieves state-of-the-art performance in terms of key IDF1, Mostly Tracked, Mostly Lost, and ID switches. We note that these are the key measures of data association, which proves that our model excels at this task in the particularly crowded scenarios present in the HiEve dataset. We also note that our MOTA is below that of SiamMOT. This can be explained due to the fact that SiamMOT makes use of a heavy-weight Faster R-CNN detector based on DLA-169~\cite{dla} that is trained on an additional large-scale dataset, CrowdHuman\cite{crowdhuman}, and therefore can find a better tradeoff between false positives and false negatives. This leads to an improved MOTA, despite having higher ID Switches. Instead, we use Tracktor for preprocessing with a ResNet50 backbone, and we use exclusively the provided training data in HiEve. 

\subsubsection{Multi-Object Tracking and Segmentation}
\noindent{\bf Mask R-CNN MOTS Baselines}.
Following a similar approach to \cite{voigtlaender2019mots}, we take several top-performing MOT methods that use public detections on the MOT17 benchmark and generate segmentation masks on their precomputed outputs with a MaskR-CNN trained on COCO that shares the same backbone with ${\tilde{N}}_v^{enc}$. We report the results on the MOTS20 training set in Table \ref{tab:mot17mots}. For a fair comparison, we run our model with MOT17 public detections in this experiment. Note that we use a stronger segmentation model with the baselines compared to \cite{voigtlaender2019mots}, and we improve on the original scores reported by them.  It is worth highlighting that we report our 4-fold cross-validation results, whereas the baselines are already trained on these sequences together with additional sequences from the MOT17 training set. Remarkably, despite this disadvantage, our method outperforms the baselines. These results provide strong evidence that our unified approach that performs joint tracking and segmentation can improve over methods that detach segmentation from tracking. 

%!TEX root = ../1404.tex

\begin{table*}
\center
\tabcolsep=0.11cm

    \resizebox{\columnwidth*2}{!}{
    \begin{tabular}{l c c c c c c c c c c }
     \toprule
     Method & Val. & sMOTSA $\uparrow$ & IDF1 $\uparrow$ & MOTSA $\uparrow$  & MT $\uparrow$ & ML $\downarrow$ & FP $\downarrow$ & FN $\downarrow$ & ID Sw. $\downarrow$ \\ [0.5ex] 
    \midrule
     \multicolumn{10}{c}{MOTS20 Train Set - Public} \\
     \midrule
     jCC~\cite{8493320} & \xmark & 48.8 & 61.6 & 63.3 &  81 & 32 & 2514 & 7159 & 202 \\
    MHT\_DAM~\cite{Kim_2015_ICCV} & \xmark & 51.4 & 62.6 & 65.4 &  82 & 34 & 2266 & 6877 & 164 \\
     FWT~\cite{Henschel_2018_CVPR_Workshops} & \xmark & 51.5 & 54.0 & 65.4 & 81 & 33 & 1835 & 7213 & 249 \\
    Lif\_T~\cite{pmlr-v119-hornakova20a} & \xmark & 53.6 & \textbf{73.3} & 67.5 & \textbf{89} & 23 & 2026 & \textbf{6615} & \textbf{92} \\
    CenterTrack~\cite{10.1007/978-3-030-58548-8_28} & \xmark & 53.9 & 58.4 & 67.5 &  80 & 34 & 1488 & 6975 & 279 \\
    Tracktor++~\cite{tracktor} & \xmark & 54.6 & 62.7 & 67.7 &  72 & 24 & 1036 & 7509 & 154 \\
    \MPNSeg (Ours) & \cmark & \textbf{55.4} & 68.2 & \textbf{67.8} & 76 & 39 & \textbf{955} & 7592 & 102 \\
     \bottomrule
    \end{tabular}}

\caption{Comparison of our method with the baselines obtained from MOT methods on \textbf{MOTS20 train} set. We report our \textit{cross-validation} scores, whereas baselines are obtained from the corresponding method's \textit{training set} results.}
\vspace{-0.2cm}
\label{tab:mot17mots}

\end{table*}

\noindent{\bf MOTS20}. We present our results on MOTS20 train and test sets in Table \ref{tab:motschallenge}. %\lau{table num wrong} \orc{\checkmark} 
For a fair comparison on the training set, we follow TrackFormer's 4-fold cross-validation scheme. In this setting, we observe 1.2 sMOTSA and 3.7 MOTSA improvements over TrackFormer, and 1.8 and 4.9 MOTSA improvements over PointTrack. On the MOTS20, test set, our model establishes a new state-of-the-art by significantly improving on all metrics over the top published methods. Specifically, we outperform TrackFormer, TraDeS,  and TrackR-CNN by 3.7, 7.8, and 18.2 points in sMOTSA and 5.2, 10.1, and 16.4 points in IDF1, respectively. In addition, our model achieves the highest track coverage among all methods and reduces the number of identity switches by 25\% compared to the closest method.  
%!TEX root = ../1404.tex

\begin{table*}
\center
\tabcolsep=0.11cm

    \resizebox{\columnwidth*2}{!}{
    \begin{tabular}{l c c c c c c c c c }
     \toprule
     Method & sMOTSA $\uparrow$ & IDF1 $\uparrow$ & MOTSA $\uparrow$  & MT $\uparrow$ & ML $\downarrow$ & FP $\downarrow$ & FN $\downarrow$ & ID Sw. $\downarrow$ \\ [0.5ex] 
     
    \midrule
     \multicolumn{9}{c}{MOTS20 Train Set} \\
     \midrule

          TrackR-CNN~\cite{voigtlaender2019mots} & 52.7 & -- & 66.9 &  -- & -- & -- & -- & -- \\
          MOTSNet~\cite{porzi2020learning} & 56.8 & -- & 69.4 &  -- & -- & -- & -- & -- \\
          PointTrack~\cite{xu2020segment}  & 58.1 & -- & 70.6 &  -- & -- & -- & -- & -- \\
          TrackFormer~\cite{meinhardt2021trackformer} &  58.7 & -- & -- &  -- & -- & -- & -- & -- \\
          \MPNSeg (Ours)  & \textbf{59.9} & 71.9 & \textbf{74.3} & 126 & 17 & 1679 & 5017 & 210 \\
    \midrule
     \multicolumn{9}{c}{MOTS20 Test Set} \\
     \midrule

     TrackR-CNN~\cite{voigtlaender2019mots} & 40.6 & 42.4 & 55.2 &  127 & 71 & 1261 & 12641 & 567 \\
          TraDeS~\cite{Wu_2021_CVPR} & 50.8 & 58.7 & 65.5 &  162 & 60 & 1474 & 9169 & 492 \\
          TrackFormer~\cite{meinhardt2021trackformer} & 54.9 & 63.6 & -- &  -- & -- & 2233 &  \textbf{7195} & 278 \\
     \MPNSeg (Ours) & \textbf{58.6} &  \textbf{68.8} &  \textbf{73.7} &   \textbf{207} &  \textbf{26} &  \textbf{1059} & 7233 &  \textbf{202} \\

     \bottomrule
    \end{tabular}}

\caption{Comparison of our method with state-of-the-art on \textbf{MOTS20 train and test} sets.}
\vspace{-0.2cm}
\label{tab:motschallenge}

\end{table*}

\noindent{\bf KITTI MOTS}. We compare our method against the top published methods on KITTIMOTS in Table \ref{tab:kittimots-wide}. We achieve the best performance among the methods that only utilize 2D information and surpass the previous state-of-the-art, PointTrack, by 1.1 HOTA. Our approach excels in association accuracy (AssA) and MT, in which we outperform the closest method with 4 points and 20\% more track coverage. In fact, our 2D approach obtains highly competitive results even when compared to the methods that make use of 3D information, such as EagerMOT and MOTSFusion. Specifically, we obtain a significantly higher MT while also preserving 50\% more tracks. Furthermore, our model produces the fewest number of identity switches among all 2D and 3D methods. 

%!TEX root = ../1404.tex

\begin{table*}
\center
\tabcolsep=0.11cm

    \resizebox{\columnwidth*2}{!}{
    \begin{tabular}{l c c c c c c c c c}
     \toprule
     Method & HOTA $\uparrow$ & DetA $\uparrow$ & AssA $\uparrow$ & sMOTSA $\uparrow$ & MT $\uparrow$ & ML $\downarrow$ & FP $\downarrow$ & FN $\downarrow$ & ID Sw. $\downarrow$ \\ [0.5ex] 
     \midrule
     \multicolumn{10}{c}{2D Methods} \\
     \midrule

     TrackR-CNN~\shortcite{voigtlaender2019mots} & 41.9 &  53.8 & 33.8  & 47.3 & 123  &  36 & 5355 & 1171 & 482 \\
    PointTrack~\shortcite{xu2020segment} & 54.4 &  \textbf{62.3} &  48.0  & \textbf{61.5} &  132  & \textbf{25} & 4341  & \textbf{344} &  176 \\
      \MPNSeg (Ours) & \textbf{55.5} &  60.5 &  \textbf{52.0}  & 57.3 & \textbf{152}  & 26 & \textbf{3743} & 857 & \textbf{162} \\

    \midrule
     \multicolumn{10}{c}{3D Methods} \\
     \midrule

     MOTSFusion~\shortcite{luiten2020track} & 54.0 &  60.8 &  49.5  & 58.8 & 128  &  42 & 4868 & 463 & 279 \\
     
      EagerMOT~\shortcite{Kim21ICRA} & 57.7 &  60.3 &  56.2  & 58.1 & 117  &  37 & 5056 & \textbf{458} & 270 \\
      VIP-DeepLab~\shortcite{qiao2021vip} & \textbf{64.3} &  \textbf{70.7} &  \textbf{59.5}  & \textbf{68.8} & \textbf{199}  & \textbf{7} & \textbf{2265} & 731 & \textbf{209} \\

     \bottomrule
    \end{tabular}}

\caption{Comparison of our method with state-of-the-art on \textbf{KITTI MOTS test} set.
}
\vspace{-0.2cm}
\label{tab:kittimots-wide}

\end{table*}

\section{Discussion}
Overall, we have demonstrated strong results in both box-based and mask-based tracking in a variety of datasets. Given our graph-based formulation, our method's strongest ability is data association, as it can be observed from its high IDF1 and AssA scores, together with its fast runtime (Table \ref{tab:mot}). Moreover, as shown in Tables \ref{tab:mot17mots}, \ref{tab:motschallenge} and \ref{tab:kittimots-wide}, our method achieves top scores with our proposed attentive message passing module on MOTS benchmarks by blending tracking and segmentation cues, demonstrating the potential of solving complementary tasks jointly within a unified framework. %\gui{great explanation}

We also note, however, that our method falls short in overall tracking performance when compared to some of the more recent tracking methods that either use more complex optimization schemes \cite{pmlr-v119-hornakova20a}, or are regression-based \cite{tmoh}. Regarding the former, we believe that our method's significantly improved runtime offers a favorable efficiency tradeoff. Moreover, we note that LPC\_MOT~\cite{lpcmot} has been inspired by our work to use a message passing neural network within a multiple-hypothesis graph formulation.

Regarding regression-based approaches, we would like to note that they offer a different performance profile. Our method, since it is graph-based, focuses entirely on data association. Regression-based methods, instead, have the ability to track an increased number of boxes, and hence can improve MOTA significantly, but fall short in terms of data association performance, as can be seen from the relatively lower IDF1 scores of CenterTrack~\cite{10.1007/978-3-030-58548-8_28} and TMOH~\cite{tmoh}. We believe there is a need to investigate how these two lines of work can be combined to get the benefits of both of them. We hope future work will be able to develop such integration.

\section{Conclusion}
We have introduced a fully differentiable framework based on message passing networks that can exploit the underlying graph structure of the tracking problem. Our unified architecture reasons over the entire graph and performs data association and segmentation jointly by merging appearance, geometry, and mask features. Our approach achieves state-of-the-art results in both multi-object tracking and segmentation tasks on several benchmarks. We hope that our approach will open the door for future efforts in marrying graph-based methods with deep learning approaches and exploiting synergies between tracking and segmentation.
\\
\\

\section*{Declarations}
\noindent{\textbf{Funding.}}
This project was partially funded by the Sofja Kovalevskaja Award of the Humboldt Foundation and by the German Federal Ministry of Education and Research (BMBF) under Grant No. 01IS18036B. The authors of this work take full responsibility for its content. \\

\noindent{\textbf{Conflicts of interest.}}  Not applicable. \\

\noindent{\textbf{Availability of data and material.}} All datasets used are publicly available.\\

\noindent{\textbf{Code availability.}} Our code is publicly available.\\

\Urlmuskip=0mu plus 1mu\relax
\bibliographystyle{apacite}
\bibliography{abbrev, references}   % name your BibTeX data base

\end{document}